\documentclass[10pt,twocolumn,letterpaper]{article}

\usepackage{iccv}
\usepackage{times}
\usepackage{epsfig}
\usepackage{graphicx}
\usepackage{amsmath}
\usepackage{amssymb}

\usepackage{subcaption}
\usepackage{multirow}
\usepackage{verbatim}
\usepackage{color}
\usepackage{booktabs}

\usepackage{multibib}
\newcites{latex}{Appendix References}%

\usepackage[pagebackref=true,breaklinks=true,letterpaper=true,colorlinks,bookmarks=false]{hyperref}

\iccvfinalcopy 

\def\iccvPaperID{2472} 

\ificcvfinal\fi
\begin{document}

\title{Unsupervised Image Noise Modeling with Self-Consistent GAN}

\author{Hanshu Yan\\
	National University of Singapore\\
{\tt\small hanshu.yan@u.nus.edu}
\and
Xuan Chen\\
BGI Research\\
{\tt\small chenxuan@genomics.cn}
\and
Vincent Tan\\
National University of Singapore\\
{\tt\small vtan@nus.edu.sg}
\and
Wenhan Yang\\
National University of Singapore\\
{\tt\small eleyan@nus.edu.sg}
\and
Joe Wu\\
Biomind\\
{\tt\small joe.wu@biomind.ai}
\and
Jiashi Feng\\
National University of Singapore\\
{\tt\small elefjia@nus.edu.sg}\\
}

\maketitle

\begin{abstract}
	Noise modeling lies in the heart of many image processing tasks. However, existing deep learning methods for noise modeling generally require clean and noisy image pairs for model training; these image pairs are difficult to obtain in many realistic scenarios. To ameliorate this problem, we propose a self-consistent GAN (SCGAN), that can directly extract noise maps from noisy images, thus enabling unsupervised noise modeling.  In particular, the SCGAN introduces three novel self-consistent constraints that are complementary to one another, viz.: the noise model should produce a zero response over a clean input; the noise model should return the same output when fed with a specific pure noise input; and the noise model also should re-extract a pure noise map if the map is added to a clean image. These three constraints are simple yet effective. They jointly facilitate unsupervised learning of a noise model for various noise types.  To demonstrate its wide applicability, we deploy  the SCGAN on three image processing tasks including blind image denoising, rain streak removal, and noisy image super-resolution. The results demonstrate the effectiveness and superiority of our method over the state-of-the-art methods on a variety of benchmark datasets, even though the noise types vary significantly and paired clean images are not available. 
\end{abstract}

\section{Introduction}

Image restoration and enhancement~\cite{ntire2017}, which focus on generating high-quality images from their degraded versions, are important image processing tasks and useful for many computer vision applications. Noise modeling is critical to such tasks including image denoising~\cite{bm3d, dncnn}, noisy image super resolution~(SR)~\cite{han2017dictionary, singh2014super, yuan2018unsupervised}, and rain streak removal~\cite{yang2017deep, fu2017removing}, {\em etc.}. 

Deep learning methods have achieved astounding performances on various noisy image enhancement tasks~\cite{dncnn, ircnn, fu2017removing}. Most of these methods are designed to be supervised learning-based, and assume that noisy images together with their corresponding clean versions are available. However, in many realistic applications, it is impossible or inefficient to collect a large quantity of (clean, noise) paired images. For example, for a noisy wild natural image captured via a fixed camera, it is not possible to obtain the real clean image because of the variance of the light and objects in the scene. Thus, supervised learning-based deep models are not applicable.  In contrast, it is easy to collect from the Internet unpaired clean images, which contain different contents from the given noisy images. Even though there are no image pairs, the noisy and clean images with different contents can still together reflect the domain noise. Thus, learning a model to generate the noise is feasible. 

\begin{figure}[t]
	\begin{center}
		\includegraphics[width=1\linewidth]{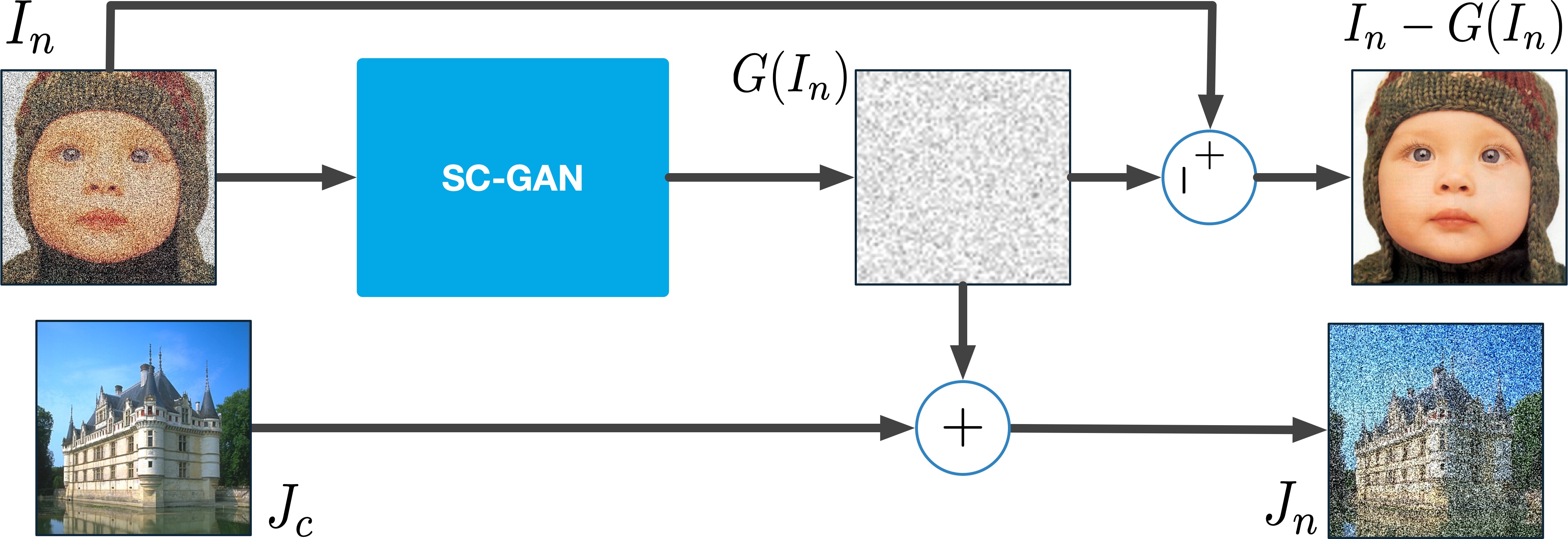}
	\end{center}
	\vspace{-1em}
	\caption{Our proposed SCGAN model learns to extract the noise map from a given noisy image $I_n$ and generates the estimated clean image. Adding the noise map to another clean image $J_c$, we can obtain $J_c$'s noisy version $J_n$, which shares similar noise patterns with $I_n$.   With the formed (noisy, clean) image pairs, like $(J_n, J_c)$, we can train a deep model in an end-to-end manner for a specific noisy image processing task. 
	}
	\label{fig:framework}
\end{figure}

By using unpaired images, our two-step method is able to address the noisy image processing problem. In the first step, also the more crucial step as shown in Fig.~\ref{fig:framework}, we learn to model the noise in the given noisy images, so that we can extract the noise maps from noisy images. Thus, image pairs can be constructed by adding the extracted noise maps to other collected clean images. In the second step,  we train a deep model with the constructed paired image set for certain task. To achieve this, we design an unsupervised deep noise modeling network, named \emph{Self-Consistent Generative Adversarial Network} (SCGAN). Given a collection of noisy images, SCGAN learns to extract their noise maps and generates their clean versions. The obtained noise model can be applied to process images with similar noise conditions. 
 
 Because of the absence of a paired dataset, the training of a GAN model is severely under-constrained. To ameliorate this, SCGAN introduces three novel self-consistent constraints for complementing the adversarial learning and facilitating the model training.
  The three constraints are developed based on the following observations. 1) A good noise model should map a clean image to the zero response; 2) A good noise model should return the same output when fed with a pure noise map. As a result, if we fix the input to be a noise map extracted from a well-trained model, the resulting output should be the same as the input noise map; 3) If we add pure noise to a clean image, a good model should extract the same noise from the resultant noisy image.  The above three constraints are complementary to one another and the adversarial loss. They are easy to be deployed in end-to-end deep model training. 

We apply the proposed SCGAN model to three noisy image processing tasks, each of which features different challenges and noise types. These three tasks include blind denoising, rain streak removal, and noisy image SR. Our unsupervised model achieves excellent performances that are even comparable with fully-supervised trained models. The main contributions of this work are as follows: (1) We propose a new architecture for unsupervised noise modeling; (2) We introduce three self-consistent losses to improve the training procedure and the performance of the model. (3) To the best of our knowledge, this is the first work to perform unsupervised rain streak removal and noisy image SR  via noise modeling. 

\begin{figure*}[t]
	\begin{center}
		\includegraphics[width=0.9\linewidth]{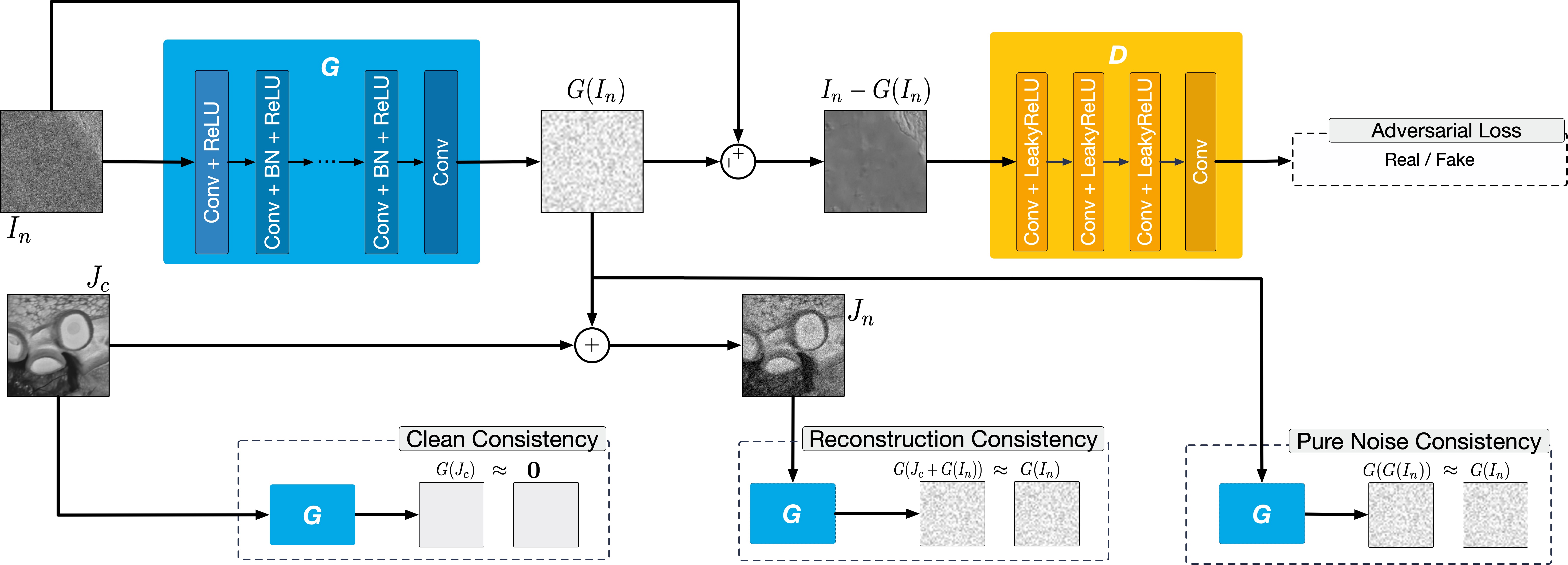}
	\end{center}
	\vspace{-1em}
	\caption{Our model consists of two networks, namely, a generator $G$ for noise map extraction and a discriminator $D$ for distinguishing real clean images from fake ones. To overcome the problem of having unpaired sets, we introduce three self-consistent losses. These include  the clean consistent loss which enforces that $G(J_c) \approx \mathbf{0}$, the pure noise consistent loss which enforces that $G(G(I_n))\approx G(I_n)$, and the reconstruction consistent loss which enforces that $G(J_c+G(I_n))\approx G(I_n)$.}
	\label{fig:arch}
\end{figure*}

\section{Related Work}

\paragraph{Deep Learning Methods for Image Restoration}
Deep learning methods have demonstrated great successes in image restoration tasks, including image denoising~\cite{dncnn,ircnn,lefkimmiatis2017non}, rain streak removal~\cite{yang2017deep, fu2017removing} and image SR~\cite{edsr, espcn, vdsr}. For the image denoising task, DnCNN~\cite{dncnn} and IRCNN~\cite{ircnn} train deep neural networks with residual blocks to learn mappings from noisy images to their corresponding residual noise maps.  For the rain streak removal task, Yang {\em et al.}~\cite{yang2017deep} and Fu {\em et al.}~\cite{fu2017removing}  proposed a multi-task deep architecture that learns the binary rain streak map to help recover clean images.  
For image SR tasks,  deep models are usually trained with high resolution (HR) images and their corresponding downsampled LR images.  EDSR~\cite{edsr} combines residual blocks and pixels shuffling in its model and achieves excellent performances on both the bicubic down-sampling and unknown degradation tracks. However, the training in all of these works is done in a supervised manner with paired noisy images and their corresponding ground truth images.  Hence, they cannot be directly used for noisy image processing tasks in the scenario that paired data is absent.
\vspace{-4mm}
\paragraph{Image Blind Denoising}
When paired data is absent, blind denoising methods are proposed to perform noise removal tasks via noise modeling techniques. Classical methods for blind image denoising usually include the noise estimation/modeling~\cite{zhu2016noise, liuxinhao, zhao2014robust, meng2013robust} and adaptive denoising procedures. Liu {\em et al.}~\cite{liuxinhao} proposed a patch-based noise level estimation method to estimate the parameters of Gaussian noise.   Zhu  {\em et al.}~\cite{zhu2016noise} used mixtures of Gaussian (MoG) to model image noises and recover clear signals with a low-rank MoG filter. These methods were developed based on human knowledge and cannot effectively handle images with unknown noise statistics. To the best of our knowledge, deep learning-based methods for noise modeling~\cite{chen2018image} are scarce. In Chen   {\em et al.}~\cite{chen2018image}, the authors proposed a smooth patch searching algorithm to obtain noise-dominant data from noisy images. These noise-dominant patches are used to train a GAN, which can model the distribution of the noise and generate noise patches. 
 However, the method requires manually tuned parameters to search for noise blocks, and the performance of noise modeling is sensitive to the parameters. Besides, the proposed noise-dominant patch search method in~\cite{chen2018image} can only model noises with zero-mean. When the mean of the additive noise, such as rain streaks, is positive,  the search method cannot extract pure rain streak patches to train a GAN for noise modeling. By contrast, in our method, we do not make any specific assumption on the additive noise. Our method does not involve searching smooth patches from noisy images, but attempts to train a deep network which directly extracts noise maps from noisy images.

\section{Approach}
\subsection{Problem Setting}
We aim to solve image noise modeling problems in an unsupervised manner. In particular, we are only provided with a set of noisy images  $\mathcal{I}_n$ and another set of clean images $\mathcal{J}_c$. The degradation of a noisy image is usually modeled as $I_n = I_c + N$~\cite{chen2018image, liuxinhao, yang2017deep}, where $I_n$ and $I_c$ represent a noisy image and its ground truth respectively, and $N$ is a noise map. We aim to obtain a noise model that can generate the accurate noise map for a noisy input image. 

\subsection{Noise Modeling with Self-Consistent GAN}
GANs have been shown to be effective in modeling complex real data distributions from a large number of sampled inputs~\cite{CGAN,DCGAN}. A GAN model consists of a generator $G$ component and a discriminator component $D$. In the noise modeling problems, we can train a GAN on the unpaired clean and noisy images to obtain a noise model $G$ that maps a noisy image $I_n$ to the noise map, i.e., $G: I_n \rightarrow N$.

In Fig.~\ref{fig:arch}, given a noisy image ${I_n}$ as the input, the noise modeling network $G$ will output the estimated noise map $G(I_n)$. The clean version of the input patch can be estimated as $I_n-G(I_n)$. By adding the extracted noise map $G(I_n)$ to some clean patch ${J_c}$, we can then generate its noisy version $J_n=J_c+G(I_n)$, which shares the same noise pattern as the given noisy patch $I_n$. The estimated clean version of a noisy input will then be sent to a discriminator $D$, together with the real clean image $J_c$.  We use an adversarial training strategy to train $G$ and $D$. 

To ensure that the training procedure is stable, the least squares loss~\cite{yuan2018unsupervised} is used as the adversarial loss. The objective function of adversarial training is:
\begin{align}
\mathcal{L}_{\mathrm{GAN}}(G,D) = & - \mathbb{E}_{J_c \in \mathcal{J}_c}\big[\|D(J_c)-\mathbf{1}\|^2_2\big]   \nonumber \\
& -  \mathbb{E}_{I_n \in \mathcal{I}_n}\big[\|D(I_n-G(I_n))-\mathbf{0}\|^2_2\big] . 
\label{eq:ganloss}
\end{align} 
We train the discriminator $D$ to maximize this objective. At the same time, we optimize  $G$  to minimize the loss such that $G$ can generate noisy maps presenting similar noise as $\mathcal{I}_n$. Namely, we consider $$\displaystyle \min_{G}\max_{D} ~ \mathcal{L}_{\mathrm{GAN}}(G,D).$$
The above optimization will achieve a Nash equilibrium and the optimized $G$ can be deployed as a noise model for new inputs directly. 

Though the above application of vanilla GAN for noise modeling is straightforward, we find that if we  only use the adversarial loss, the training of noise modeling is severely under-constrained because of the absence of paired image patches. To ameliorate this problem, we carefully study the generator $G$. We derive three implicit constraints, which we term as \emph{self-consistent} losses~(Fig.~\ref{fig:arch}), from inherent properties of the noise modeling mapping. 

\begin{enumerate}
	\item The {first} intuitive property shown in Eqn.~\eqref{eq:clean} is that, taking input as a noise-free clean image $J_c$, the generator $G$ should output zero response: 
	\begin{equation}
		\mathcal{L}_{\mathrm{clean}} = \mathbb{E}_{J_c \in \mathcal{J}_c}\big[\|G(J_c)-\mathbf{0}\|^2_2\big].
		\label{eq:clean}
	\end{equation}
	
	\item We term an image, that only contains noise sampled from the noise distribution, as a pure noise image. The {second} property is that a pure noise image, after passing through the noise model $G$,  should produce the same pure noise map. However, since pure noise images are not available in the training set, we propose another constraint (Eqn.~\eqref{eq:purenoise}) as an alternative to enforce such a desired property. When training the model, given a noisy image $I_n$, the output $G(I_n)$ will be fed back to $G$. The $G$ is optimized  to minimize the difference between $G(I_n)$ and $G(G(I_n))$ as below: 
	
	\begin{equation}
		\mathcal{L}_{\mathrm{pn}} =  \mathbb{E}_{I_n \in \mathcal{I}_n}\big[\|G(G(I_n)) - G(I_n)\|^2_2\big].
		 \label{eq:purenoise}
	\end{equation}
	
	\item The {third} property is exactly the definition of noise modeling: for any noisy image which is the addition of a clean image and a pure noise map, the pure noise map should be extracted from the noisy image correctly by $G$. As shown in Eqn.~\eqref{eq:rec}, added by a clean image $J_c$, the extracted noise $G(I_n)$ is fed back to $G$ again. $G$ should be able to reconstruct the noise map. This is dictated by the following constraint:
	\begin{equation}
		\mathcal{L}_{\mathrm{rec}}	=  \mathbb{E}_{\substack{J_c \in \mathcal{J}_c \\ I_n \in \mathcal{I}_n}}\big[\|G(J_c+G(I_n)) - G(I_n)\|^2_2\big].
		 \label{eq:rec}
	\end{equation}
	
\end{enumerate}

By incorporating the self-consistent implicit constraints in Eqns.~\eqref{eq:clean}--\eqref{eq:rec},  our overall objective is
\begin{equation}
\displaystyle \min_{G}\max_{D} ~ \mathcal{L}_{\mathrm{GAN}} + w_{1}  \mathcal{L}_{\mathrm{clean}} + w_{2}  \mathcal{L}_{\mathrm{pn}} + w_{3}  \mathcal{L}_{\mathrm{rec}},
\label{eq:totalloss}
\end{equation}
where $w_1, w_2$, and $w_3$ are non-negative weights.

\subsection{Architecture and Training Details of SCGAN}

\paragraph{Architecture } In Fig.~\ref{fig:arch}, our generator $G$ uses the same architecture as the model in~\cite{dncnn}. 
The first layer of $G$ is a convolutional  (conv) layer with ReLU activation, and the last layer is merely a conv layer. Each of the remaining $15$ layers is a unit consisting of a conv layer with batch normalization and ReLU activation. Both the numbers of input and output channels of the conv layers in the middle 15 units are set to $64$, and the kernel size is set as $3\times 3$. To ensure that the output noise maps have the same size as the inputs and to avoid artifacts along the edges, noisy input images are padded using the reflection method, and the padding number is set to $17$. The discriminator $D$ only has $4$ conv layers. The first $3$ conv layers are connected with a LeakyReLU activation function, where the negative slope is set to be $0.2$. The number of output channels of the $4$ conv layers are $64$, $128$,  $64$, and $1$ respectively. The kernel sizes are $5$, $5$, $3$, and $3$ and the strides are $2$, $2$, $1$, and  $1$.  All the padding numbers of each layer are set to $0$. 
\vspace{-4mm}
\paragraph{Training details} 

In order to well train a noise map extraction model, we divide the whole training procedure into {three} phases based on training epochs [\emph{ep1}, \emph{ep2}, \emph{ep3}] and accordingly schedule the changing of weight parameters.

In the first phase during epochs 0 to \emph{ep1}, the values of $w_1$, $w_2$ and $w_3$ are initialized as zero, thus only GAN loss is used for optimization. After several epochs of training,   $G$ can extract noise maps from the noisy input and generate estimated clean images. However, the recovered clean images usually present distortions and brightness-shifts. This is because $G$ wrongly treats the background and textures in images as noise and extracts them into the outputs. 

The second training phase, during epochs  \emph{ep1} to \emph{ep2}, is dedicated to this problem.  The values of $w_1$, $w_2$ increase to  preset values and the value of $w_3$ keeps being zero. Thus, $\mathcal{L}_{\mathrm{clean}}$ and $\mathcal{L}_{\mathrm{pn}}$ starts influencing the optimization. With the help of these two constraints, $G$  tends to extract zero noise map from the real clean images. $G$ also extracts the identity map for a pure noise image. At the end of this phase, the estimated clean images are free from distortion and retain similar brightness and contrast as the noisy input images. However, the extracted noise maps still contain distinct edges of noisy input images. 

To overcome this problem, in the third phase during epochs  \emph{ep2} to \emph{ep3}, the value of $w_3$ increases to positive and $\mathcal{L}_{\mathrm{rec}}$  is added to the objective function. $\mathcal{L}_{\mathrm{rec}}$ guarantees that a pure noise map can be reconstructed/re-extracted from the addition of any clean image and the pure noise map. Hence, the noise maps that are extracted by $G$ will be free from the influence of edges and textures of a certain image and a certain area in an image. In Section~\ref{sec:analysis}, we analyze the effectiveness of the proposed self-consistent losses. 

\subsection{Application to Noisy Image Restoration Tasks}
\begin{figure}[h!]
	\begin{center}
		\includegraphics[width=0.9\linewidth]{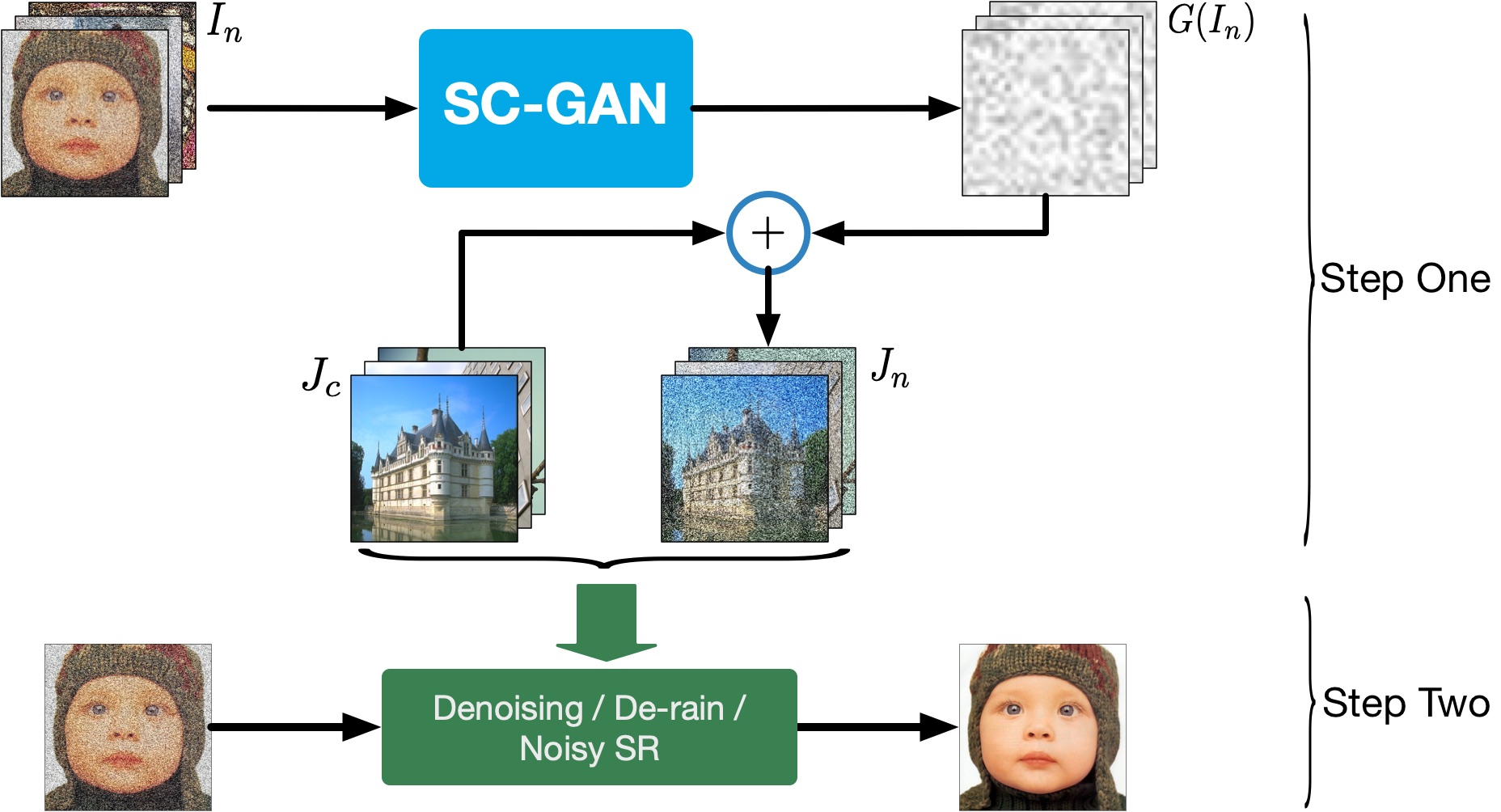}
	\end{center}
	\vspace{-1em}
	\caption{The proposed two step method for noisy image processing. In  step one, the SCGAN learns to extract noise maps from given noisy images, and add the noise maps into other clean images to construct image pairs; In step two, with the formed image pairs, a deep model for certain task can be trained  in an end-to-end manner.}
	\label{fig:app}
\end{figure}

\def \1subfig {.162} 
\def \2subfig {.85}
\begin{figure*}[h]
	\centering
	\begin{subfigure}{\1subfig \linewidth}
		\centering
		\includegraphics[width=\2subfig \linewidth]{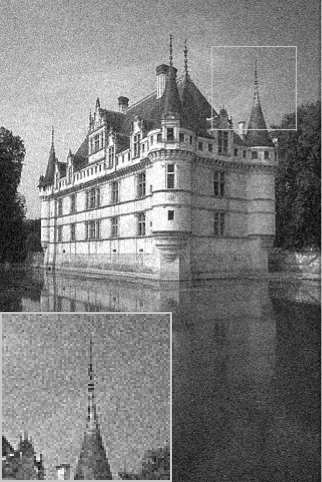}
		\caption{Noisy}
	\end{subfigure}
	\begin{subfigure}{\1subfig \linewidth}
		\centering
		\includegraphics[width=\2subfig \linewidth]{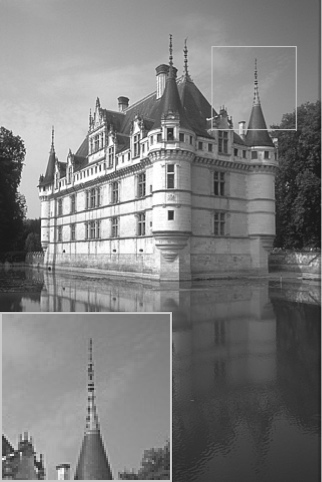}
		\caption{Ground-truth}
	\end{subfigure}
	\begin{subfigure}{\1subfig \linewidth}
		\centering
		\includegraphics[width=\2subfig \linewidth]{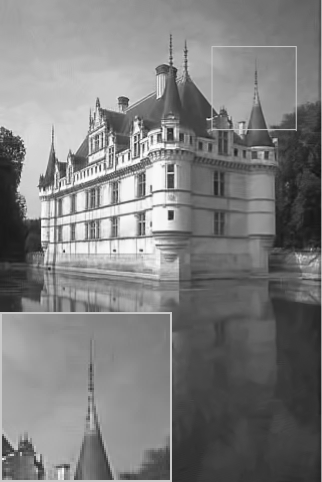}
		\caption{BM3D}
	\end{subfigure}
	\begin{subfigure}{\1subfig \linewidth}
		\centering
		\includegraphics[width=\2subfig \linewidth]{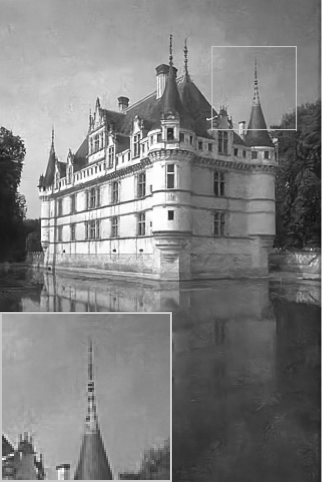}
		\caption{GCBD-2}
	\end{subfigure}
	\begin{subfigure}{\1subfig \linewidth}
		\centering
		\includegraphics[width=\2subfig \linewidth]{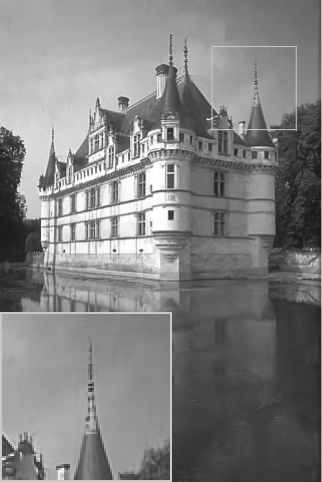}
		\caption{\textbf{\textbf{SCGAN-2}}}
	\end{subfigure}
	\begin{subfigure}{\1subfig \linewidth}
		\centering
		\includegraphics[width=\2subfig \linewidth]{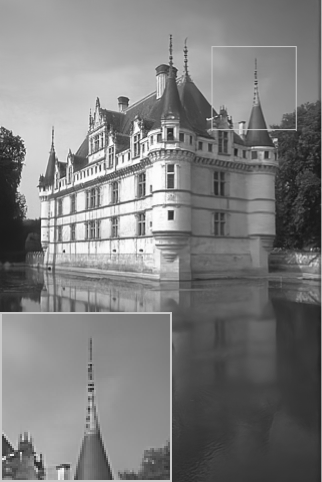}
		\caption{DnCNN-B(Oracle)}
	\end{subfigure}
	
	\caption{Comparison of denoising performance of different methods for the Image `003' from BSD68; noise level $\sigma=15$.}
	\label{fig:dn-set68}
\end{figure*}

\begin{table*}[t]
	\caption{The average PSNR (in dB) results of Gaussian noise removal on the BSD68 dataset. The BM3D and WNNM are two non-blind denoising methods, while rest are blind denoising methods.}
	\centering
	\scalebox{0.8}{
		\begin{tabular}{c|ccccccc}
			\toprule
			$\sigma$ &BM3D  & WNNM & GCBD-2 & \textbf{SCGAN-2}  & GCBD-1 & \textbf{SCGAN-1}  & DnCNN-B (Oracle) \\
			\midrule
			15 & 31.07  & 31.37 & 30.59 & 30.80 & 31.35 & 31.48  & 31.61 \\
			25 & 28.57  & 28.83 & 27.66 & 28.92  & 29.04 & 29.02 & 29.16 \\
			\bottomrule
		\end{tabular}
	}
	\label{tb:gaussian}
\end{table*}

 We apply the proposed SCGAN for blind image denoising, rain streak removal and noisy image SR~(Fig.~\ref{fig:app}). For the first two tasks, the available training data only contains a set of noisy images $\mathcal{I}_n$ with unknown noise statistics and a  set of different clean images $\mathcal{J}_c$. Using a well-trained generator $G$, we can extract noise maps from $\mathcal{I}_n$, obtain the noisy version of $\mathcal{J}_c$ and construct a paired training dataset $\lbrace \mathcal{J}_n,\mathcal{J}_c\rbrace$. We trained a DnCNN~\cite{dncnn} model for denoising and a Deep Detail Network~\cite{fu2017removing} for rain streak removal.

For the noisy image SR task, we only have a set of noisy LR images $\mathcal{I}_n$ and a set of clean HR images $\mathcal{J}_{\mathrm{HR}}$. We assume that the up-scaling factor is $r$ and the bicubic down-sampling kernel is used. Firstly,  we  down-sample $\mathcal{J}_{\mathrm{HR}}$ by the factor $r$ to get a LR clean image set $\mathcal{J}_c$. Then, following similar procedures as those mentioned in the denoising task, we generate the noisy versions $\mathcal{J}_n$ of clean LR images and construct a paired training set $\lbrace \mathcal{J}_n, \mathcal{J}_{\mathrm{HR}} \rbrace$. 
EDSR shows impressive performances on benchmarks and is the best model for the NTIRE2017 Super-Resolution Challenge~\cite{ntire2017}. We finally train an EDSR-Baseline~\cite{edsr} model with the constructed $\lbrace \mathcal{J}_n, \mathcal{J}_{\mathrm{HR}} \rbrace$ set for noisy image SR task.

\section{Experiments}

\subsection{Image Blind Denosing}
\paragraph{Dataset} 
We evaluate the performance of SCGAN for blind image denoising on the BSD68~\cite{roth2005fields} benchmark dataset. The BSD68 set consists of 68 images with the resolution of $321 \times 481$, and the images cover a variety of scenes including animals, human, buildings and natural scenery.  We synthesize the noisy testing images by adding Gaussian noise to the BSD68 dataset.


\vspace{-4mm}
\paragraph{Baselines} We compare the performance of our proposed SCGAN with  state-of-the-art \emph{blind} denoising methods, including DnCNN-B~\cite{dncnn} and GCBD~\cite{chen2018image}, and classical \emph{non-blind} denoising methods such as BM3D~\cite{bm3d} and WNNM~\cite{gu2014weighted}. 
For the DnCNN-B method, it is trained with real Gaussian noisy images in a fully supervised manner, where the noisy images are synthesized by adding noises from the range of $\sigma \in [0,55]$ to a set of 400-clean images~\cite{chen2017trainable} of size $180\times180$. We regard DnCNN-B as the performance { upper bound}. 

 The GCBD method~\cite{chen2018image} shares a similar two-step framework as our method to perform blind denoising. In GCBD, noise blocks are extracted from noisy images to train a GAN for modeling the noise distribution. This is to facilitate forming a dataset of paired images to train a DnCNN model. However, the paper~\cite{chen2018image} does not provide details about the images used for noise modeling and the set of clean images for DnCNN training. Thus, for a fair comparison, we tried our best to reproduce the GCBD method and tested it on the same datasets that are used for SCGAN. 
\vspace{-4mm}

\def \1subfig {.162} 
\def \2subfig {1}
\begin{figure*}[h!]
	\centering
	\begin{subfigure}{\1subfig \linewidth}
		\centering
		\includegraphics[width=\2subfig \linewidth]{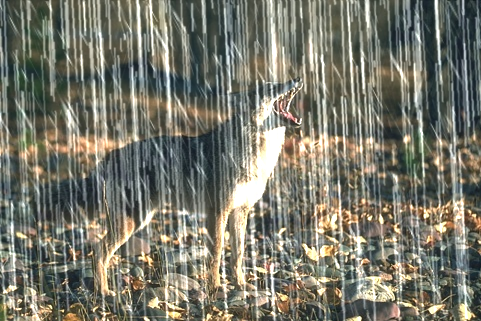}
	\end{subfigure}
	\begin{subfigure}{\1subfig \linewidth}
		\centering
		\includegraphics[width=\2subfig \linewidth]{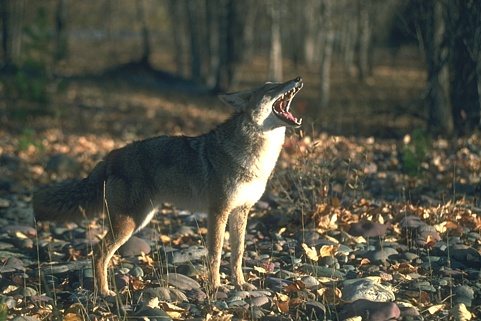}
	\end{subfigure}
	\begin{subfigure}{\1subfig \linewidth}
		\centering
		\includegraphics[width=\2subfig \linewidth]{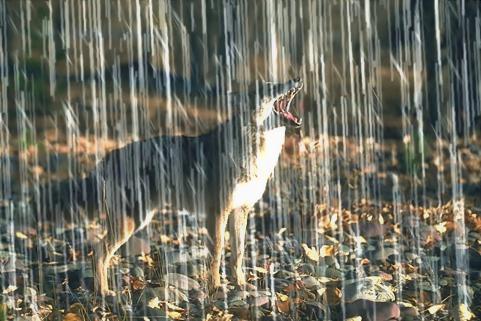}
	\end{subfigure}
	\begin{subfigure}{\1subfig \linewidth}
		\centering
		\includegraphics[width=\2subfig \linewidth]{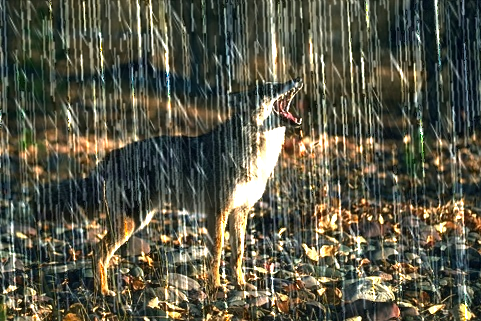}
	\end{subfigure}
	\begin{subfigure}{\1subfig \linewidth}
		\centering
		\includegraphics[width=\2subfig \linewidth]{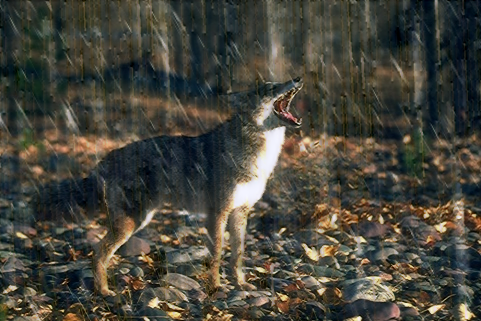}
	\end{subfigure}
	\begin{subfigure}{\1subfig \linewidth}
		\centering
		\includegraphics[width=\2subfig \linewidth]{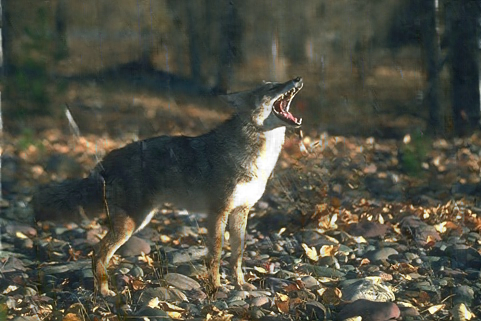}
	\end{subfigure}
	\vspace{2mm}
	
	\begin{subfigure}{\1subfig \linewidth}
		\centering
		\includegraphics[width=\2subfig \linewidth]{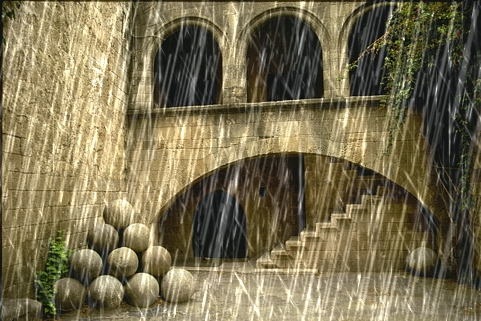}
		\caption{Rainy}
	\end{subfigure}
	\begin{subfigure}{\1subfig \linewidth}
		\centering
		\includegraphics[width=\2subfig \linewidth]{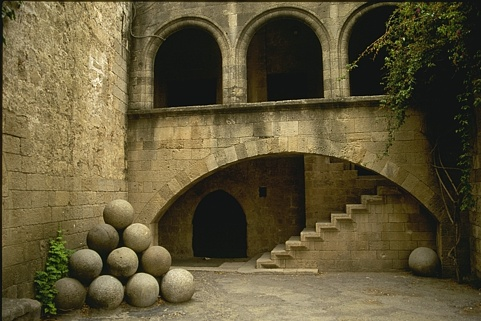}
		\caption{Ground Truth}
	\end{subfigure}
	\begin{subfigure}{\1subfig \linewidth}
		\centering
		\includegraphics[width=\2subfig \linewidth]{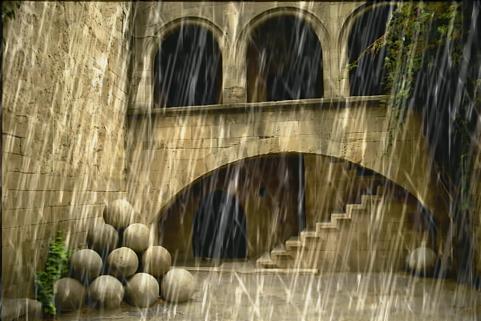}
		\caption{LP}
	\end{subfigure}
	\begin{subfigure}{\1subfig \linewidth}
		\centering
		\includegraphics[width=\2subfig \linewidth]{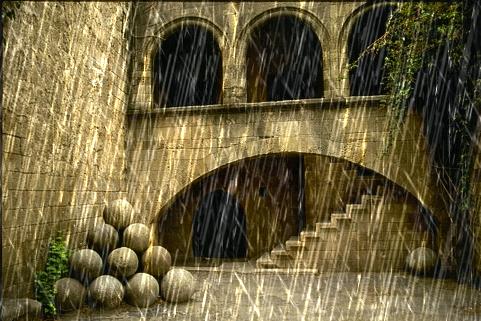}
		\caption{DSC}
	\end{subfigure}
	\begin{subfigure}{\1subfig \linewidth}
		\centering
		\includegraphics[width=\2subfig \linewidth]{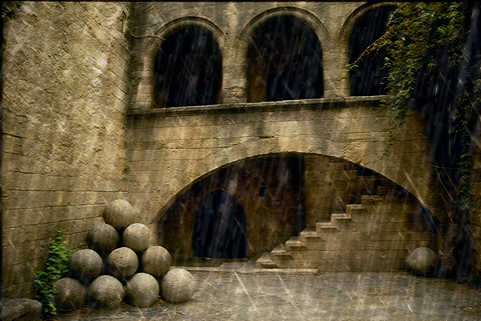}
		\caption{\textbf{SCGAN}}
	\end{subfigure}
	\begin{subfigure}{\1subfig \linewidth}
		\centering
		\includegraphics[width=\2subfig \linewidth]{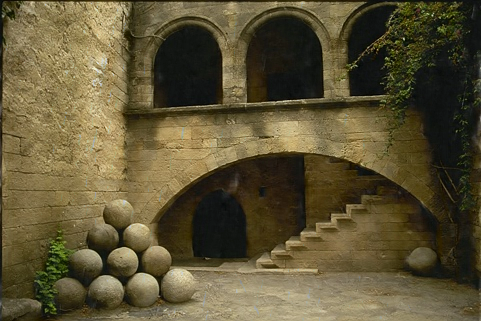}
		\caption{JORDER-}
	\end{subfigure}
	
	\caption{Comparison of de-rain performance of different methods for the  image `22' and the image `28' from the Rain100H test set. The JORDER- is a deep model trained in a fully supervised way.}
	\label{fig:derain}
\end{figure*}

\paragraph{Settings} We conducted experiments with two settings. In the first setting, the given training noisy images contain a large number of smooth patches.  Thus, we collect $200$ clean images online, and all the images collected only contain the sky or regions consisting of pure color. We divide these images into two sets: one set of images are added with Gaussian noises of certain intensities and so as to form the noisy image set $\mathcal{I}_n$; the other set of clean images is used as $\mathcal{J}_c$. We cropped the images of two sets into patches of size $128\times 128$ and subtracted their means.  

In the second setting, the available noisy images for training are constructed by adding Gaussian noise to clean images in the DIV2K~\cite{ntire2017} dataset.  This dataset consists of a diverse set of images, such as people, natural scenes, and handmade objects. We downsampled the images $1$ to $800$ by a factor of $2$ and divided them into two sets, indexing from 1-400 and 401-800 respectively. Next, we added Gaussian noises to the first set and formed the noisy image set $\mathcal{I}_n$. The other set is used as the clean image set $\mathcal{J}_c$.

In both two settings, we firstly train an SCGAN model to extract noise maps from the set $\mathcal{I}_n$. For fair comparisons, in the second step, we trained our DnCNN denoising model on the same dataset as that used in DnCNN-B; thus we added noise maps extracted from $\mathcal{I}_n$ into the 400 clean images~\cite{chen2017trainable} to construct the paired training set. The parameter settings for training DnCNN are the same as those in~\cite{dncnn}.
\vspace{-4mm}
\paragraph{Results}
For the first setting, we conducted experiments with $\mathcal{I}_n$ and $\mathcal{J}_c$ that are formed with the collected 200 images. 
With the generated paired data, we trained a DnCNN model and tested it on the BSD68 set. The results of SCGAN (SCGAN-1) and GCBD (GCBD-1) are listed in Table~\ref{tb:gaussian}. It is observed that our method is comparable to GCBD. For $\sigma=15$, the PSNR of SCGAN-1 is $31.48$dB, higher than GCBD-1's $31.35$dB, and close to the upper bound (using DnCNN-B) of $31.61$dB. For $\sigma=25$, our method is worse than GCDB-1 by $0.02$dB, which is marginal. However, it is still close to the upper bound and better than two non-blind methods, BM3D and WNNM. 

For the second setting, we conducted experiments with $\mathcal{I}_n$ and $\mathcal{J}_c$  that are constructed with images from the DIV2K set. The visual results are shown in Fig.~\ref{fig:dn-set68}, and
the PSNR value of our method~(SCGAN-2) is still close to the upper bound.  For $\sigma=25$, SCGAN-2 is better than two non-blind methods. To compare with GCBD, we reproduced the GCBD method (GCBD-2) using the same parameter settings in~\cite{chen2018image} for noise blocks extraction from the DIV2K dataset. For $\sigma=15$, the result of GCBD-2 is slightly lower than SCGAN-2. However, for $\sigma=25$, SCGAN outperforms GCBD by almost  $1.3$dB. In the GCBD method, the training of the GAN for noise modeling is sensitive to similarities between extracted noise blocks and real Gaussian noises. However, we observed that the extracted noise blocks from DIV2K mostly contain textures related to the image contents. The manually set parameters result in extracted noise blocks that may not be similar to the real noise.  Comparatively, our proposed self-consistent losses can help SCGAN extract noise maps even from non-smooth areas as analyzed in Section~\ref{sec:analysis}. 

\subsection{Rain streak Removal}

To show that the proposed SCGAN can model more complex noise, we apply it to a rain streak removal task.  The degradation of rain streaks can be model as $I_n=I_c+N$~\cite{huang2012context}, where the rainy image $I_n$ is regarded as the additive combination of its ground truth $I_c$ with a rain streak map $N$. The mean value of the rain streak map is usually positive. Thus, the smooth patch search method in GCBD~\cite{chen2018image} cannot be applied, as it violates the critical assumption of zero-mean noise that is used by the GCBD. 
\vspace{-4mm}
\paragraph{Dataset}
We evaluate the proposed SCGAN model on the Rain100H dataset~\cite{yang2017deep}, which consists of 1800 $(\mathrm{rainy},\mathrm{clean})$ pairs in the training set and 200 pairs in the test set. The 1800 rainy images in the training set are used to form the given noise image set $\mathcal{I}_n$. We select 900 clean images from DIV2K dataset, where all the clean images have different contents from the rainy images. The selected 900 clean images form the clean image set $\mathcal{J}_c$. 

\begin{table}[h]
	\caption{The average PSNR (in dB) results of rain streak removal on the Rain100H dataset. The JORDER- is a deep model trained in a fully supervised way.}
	\centering
	\footnotesize
	\scalebox{0.95}
	{
		\begin{tabular}{c|cccccc}
			\toprule
			Methods & ID~\cite{kang2012automatic}   & LP~\cite{li2016rain} & DSC~\cite{luo2015removing} & \textbf{\textbf{SCGAN}} & JORDER-~\cite{yang2017deep} \\
			\midrule
			PSNR & 14.02 & 14.26 & 15.66 & 19.865 & 20.79 \\
			\bottomrule
		\end{tabular}
	}
	\label{tb:derain}
\end{table}

\vspace{-4mm}
\paragraph{Baselines}
We train an SCGAN model with $\mathcal{I}_n$ and $\mathcal{J}_c$, and finally, obtain a model that can efficiently extract rain streak maps from given rainy images. To construct a paired training set, the extracted rain streak maps are added to clean images in $\mathcal{J}_c$. With the image pairs, we further train a deep detailed network~(DDN) for rain streak removal in an end-to-end way. The architecture of de-rain model follows the that in~\cite{fu2017removing}. 
We compare our method with several classical methods, such as an image decomposition method~(ID)~\cite{kang2012automatic}, a layer prior method~(LP)~\cite{li2016rain}  and a sparse coding method~(DSC)~\cite{luo2015removing}. Besides, we also compare our method with a deep learning method, JORDER-~\cite{yang2017deep}, which is trained in a fully supervised way.
\vspace{-4mm}
\paragraph{Results}
The result of our SCGAN method is shown in Table~\ref{tb:derain} and Fig.~\ref{fig:derain}. As shown in Fig.~\ref{fig:derain}, the SCGAN can efficiently remove the rain streaks. The visual result of SCGAN is much better than three classical methods, i.e., ID, LP and DSC, and close to the fully-supervised method, JORDER-.  Quantitatively, the SCGAN distinctly outperforms three classical methods. It is 5.8dB better than the ID~\cite{kang2012automatic}, 5.6dB better than the LP~\cite{li2016rain} and 4.2dB better than the DSC~\cite{luo2015removing}. Besides, we also compare our method with the fully supervised JORDER- that is trained with the whole Rain100H training set, and the performance of SCGAN is comparable to that of JORDER-~\cite{yang2017deep}. Our results suggest that the rain streak maps extracted by SCGAN are similar to the real rain streaks contained in the Rain100H set.

\def \1subfig {.162} 
\def \2subfig {1}
\begin{figure*}[t]

	\centering
	\begin{subfigure}{\1subfig \linewidth}
		\centering
		\includegraphics[width=\2subfig \linewidth]{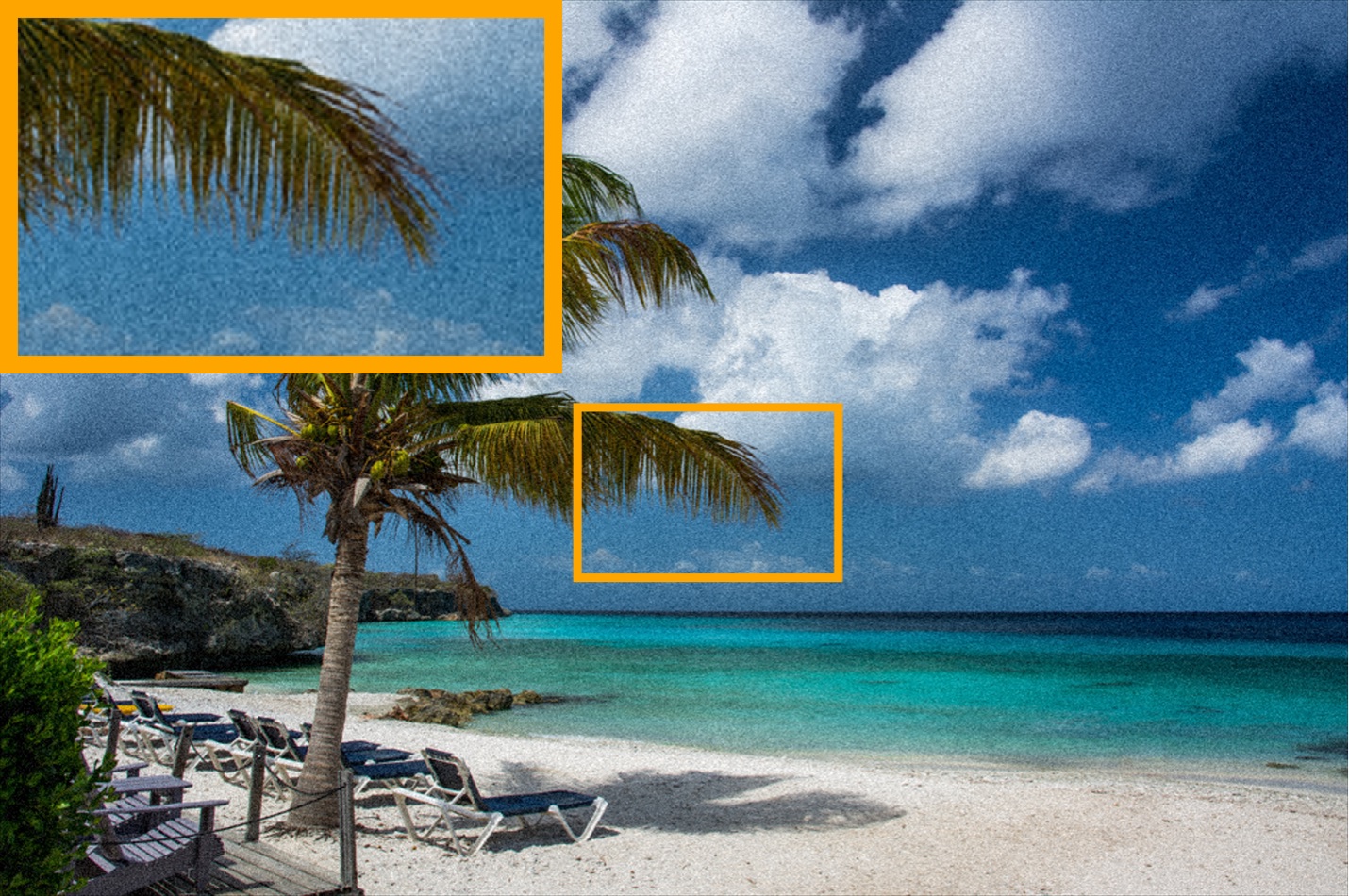}
	\end{subfigure}
	\begin{subfigure}{\1subfig \linewidth}
		\centering
		\includegraphics[width=\2subfig \linewidth]{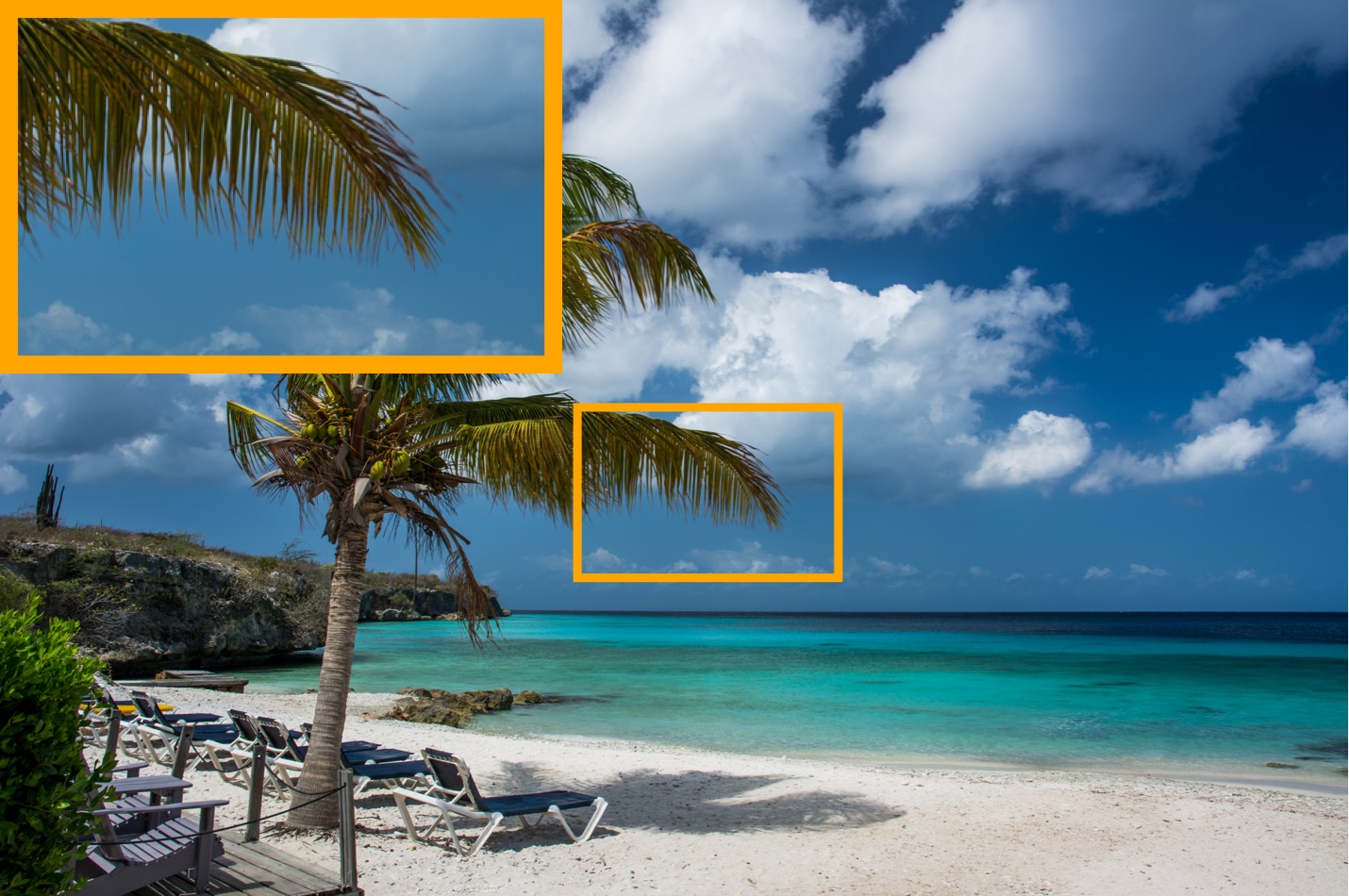}
	\end{subfigure}
	\begin{subfigure}{\1subfig \linewidth}
		\centering
		\includegraphics[width=\2subfig \linewidth]{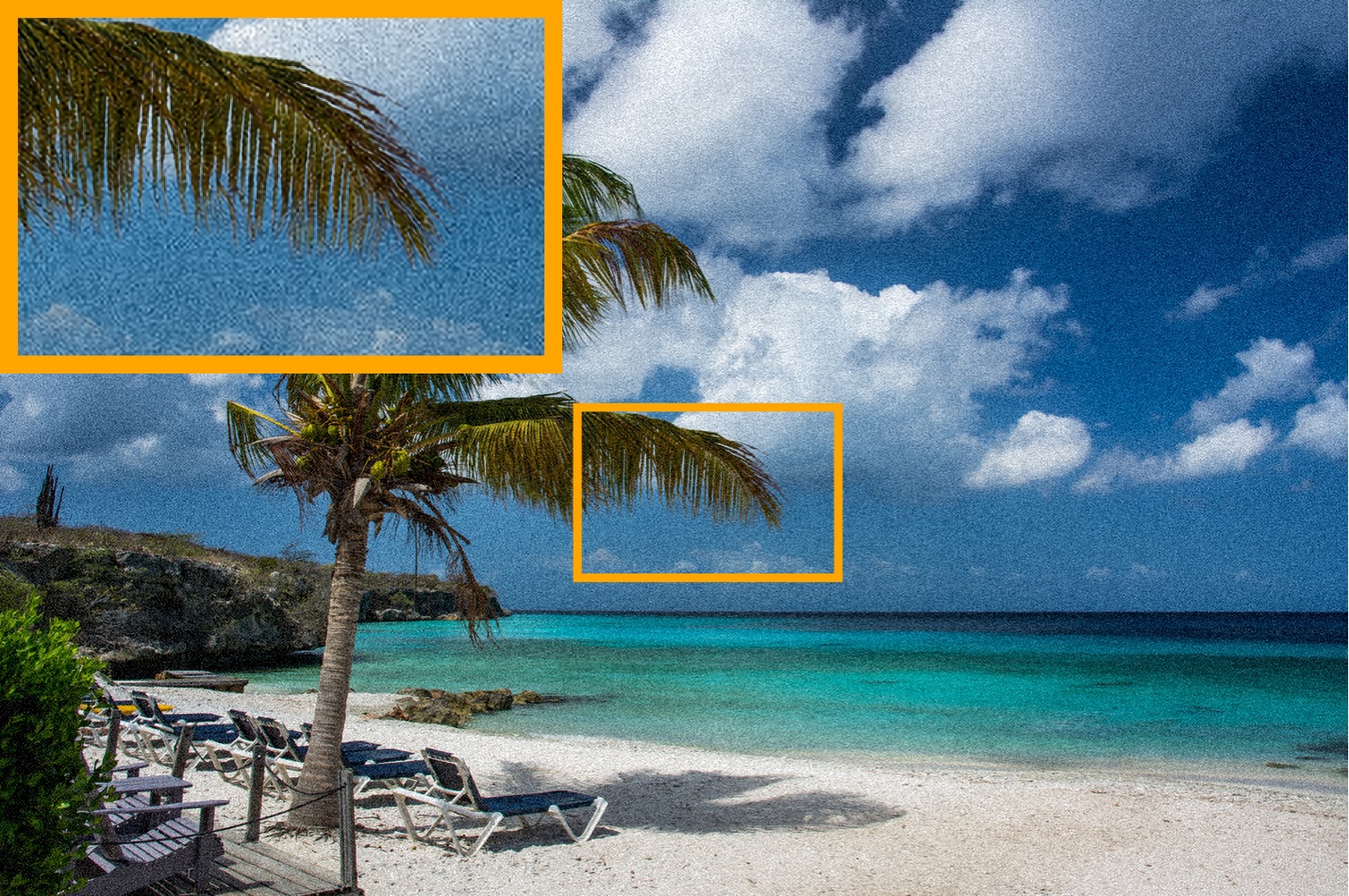}
	\end{subfigure}
	\begin{subfigure}{\1subfig \linewidth}
		\centering
		\includegraphics[width=\2subfig \linewidth]{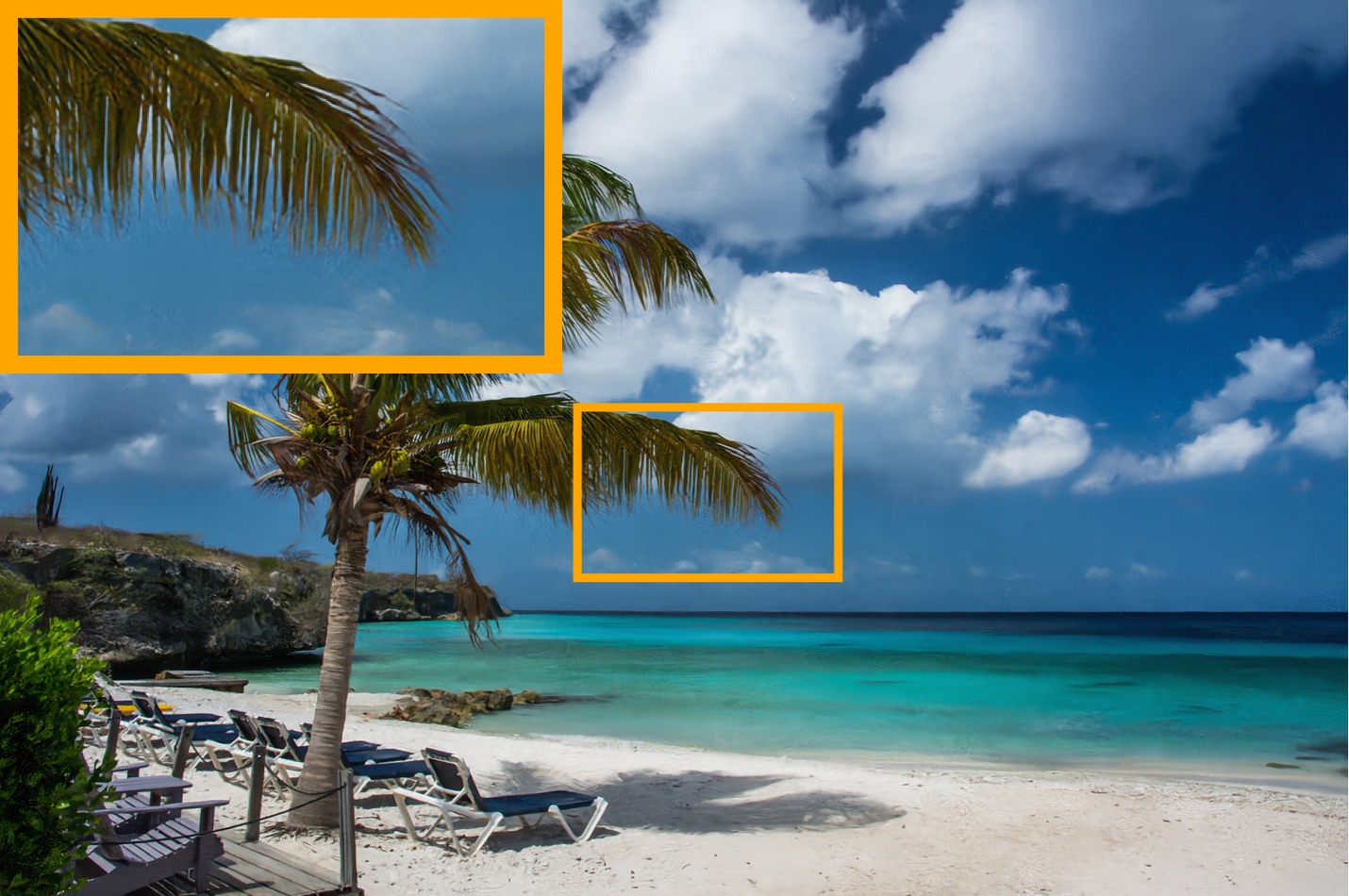}
	\end{subfigure}
	\begin{subfigure}{\1subfig \linewidth}
		\centering
		\includegraphics[width=\2subfig \linewidth]{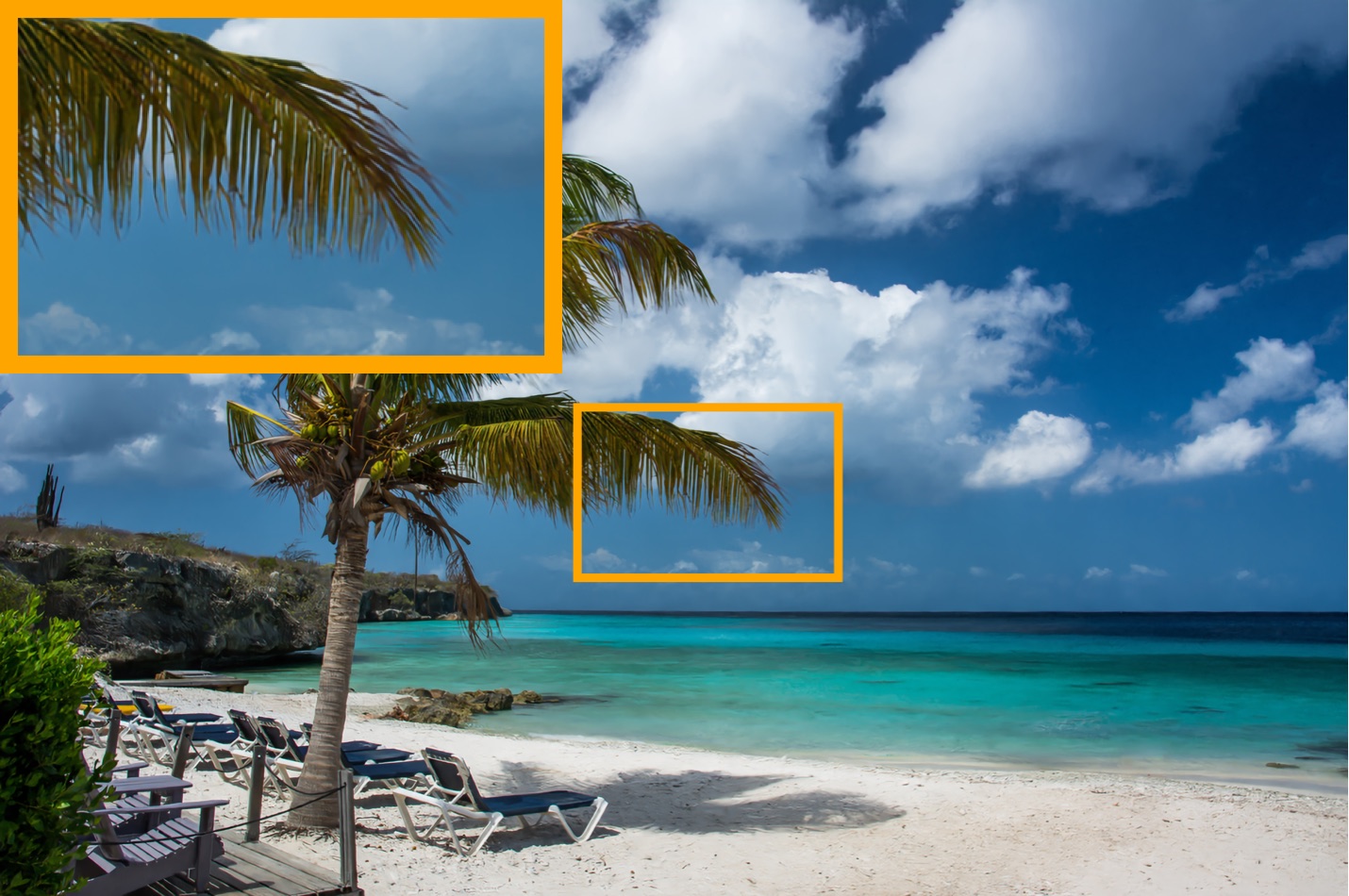}
	\end{subfigure}
	\begin{subfigure}{\1subfig \linewidth}
		\centering
		\includegraphics[width=\2subfig \linewidth]{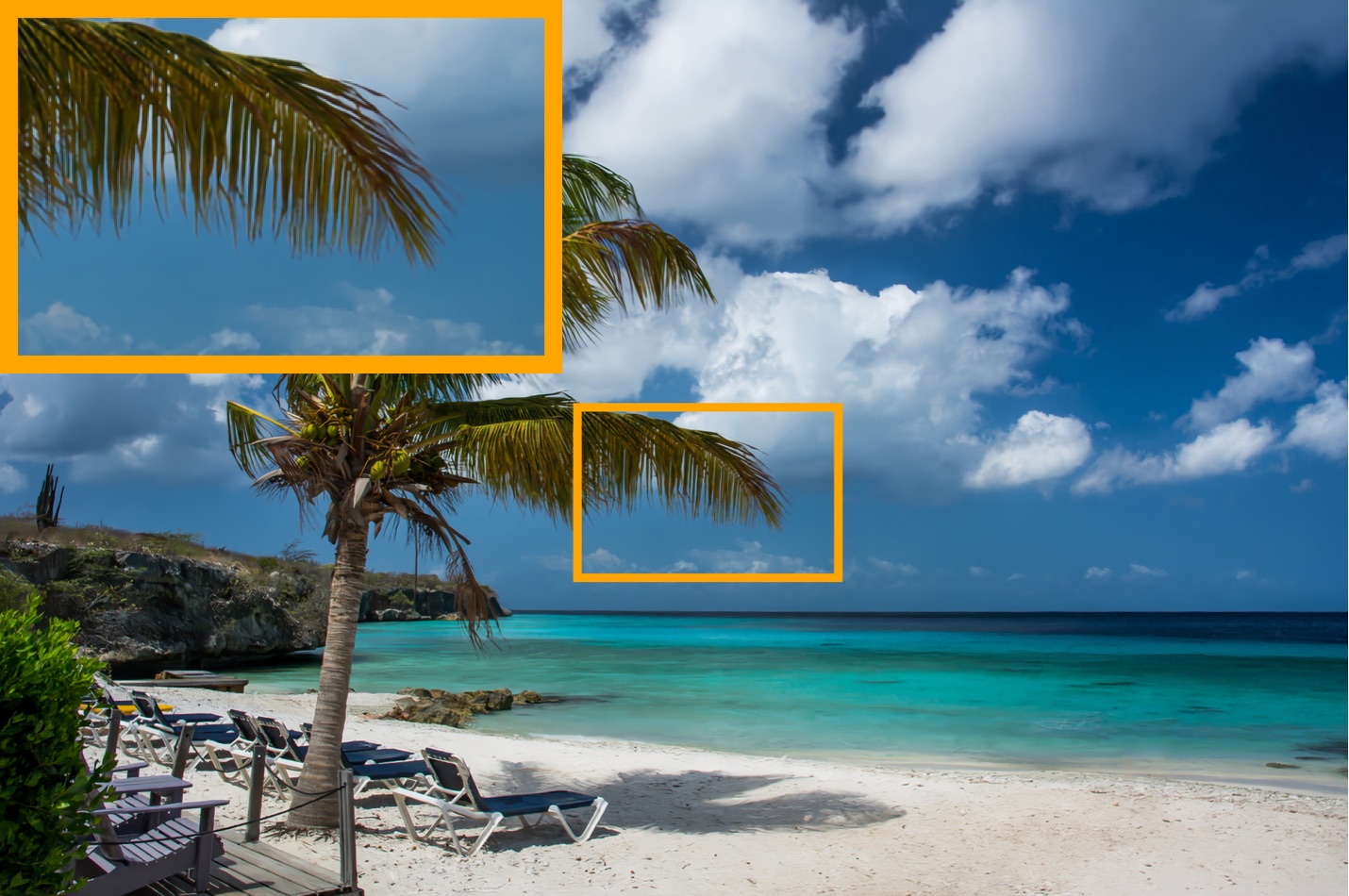}
	\end{subfigure}
	
	\vspace{2mm}
	
	\centering
	\setcounter{subfigure}{0}
	\begin{subfigure}{\1subfig \linewidth}
		\centering
		\includegraphics[width=\2subfig \linewidth]{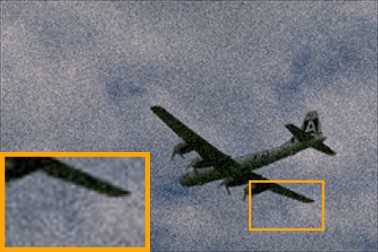}
		\caption{Bicubic}
	\end{subfigure}
	\begin{subfigure}{\1subfig \linewidth}
		\centering
		\includegraphics[width=\2subfig \linewidth]{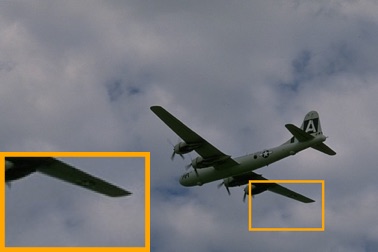}
		\caption{Ground truth}
	\end{subfigure}
	\begin{subfigure}{\1subfig \linewidth}
		\centering
		\includegraphics[width=\2subfig \linewidth]{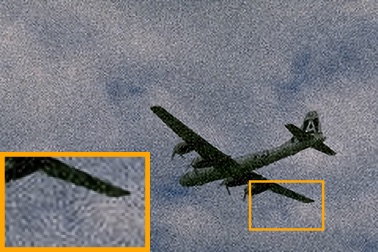}
		\caption{EDSR-b}
	\end{subfigure}
	\begin{subfigure}{\1subfig \linewidth}
		\centering
		\includegraphics[width=\2subfig \linewidth]{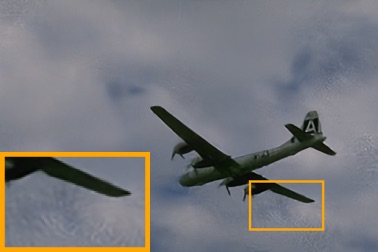}
		\caption{Dn+SR}
	\end{subfigure}
	\begin{subfigure}{\1subfig \linewidth}
		\centering
		\includegraphics[width=\2subfig \linewidth]{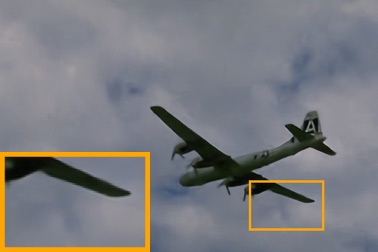}
		\caption{\textbf{\textbf{SCGAN}}}
	\end{subfigure}
	\begin{subfigure}{\1subfig \linewidth}
		\centering
		\includegraphics[width=\2subfig \linewidth]{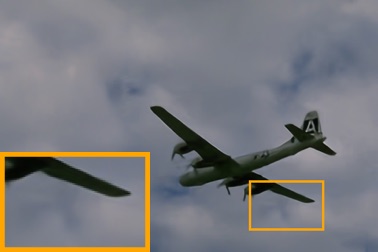}
		\caption{EDSR* (Oracle)}
	\end{subfigure}

	\caption{From top to bottom, SR visual results of an image (0806) from the DIV2K dataset and one image (3096) from the BSD100 dataset with upscaling factor $\times2$ are shown. LR noisy images contain  Gaussian noises with $\sigma=10$.}
	\label{fig:sr}
	
\end{figure*}

\subsection{Noisy Image Super-Resolution}
\paragraph{Dataset}
The third task we evaluate the SCGAN for noise modeling is the super-resolution over Gaussian noisy images. We evaluated the model on two benchmark datasets, namely, the DIV2K~\cite{ntire2017} validation set and the BSD100~\cite{B100} set, and the Gaussian noise is added to these two datasets.

The unpaired image sets used for training SCGAN are formed as the following: The DIV2K training set consists of 800 HR clean images. We downsampled them by applying bicubic interpolation and divided the obtained LR images into two sets, one of which consists of images indexed from $1$ to $400$ and the other consists of images indexed from $401$ to $800$. We then added Gaussian noises to the first set and kept the other set unchanged. Hence,    a noisy image set $\mathcal{I}_n$ and a clean image set $\mathcal{J}_c$ are obtained. 

In this experiment, only images in $\mathcal{J}_c$ have corresponding HR images $\mathcal{J}_{\mathrm{HR}}$. With the sufficiently well-trained SCGAN, we can synthesize a noisy image set $\mathcal{J}_n$ by adding noises extracted from $\mathcal{I}_n$ to $\mathcal{J}_c$. Finally, we obtain a paired image set $\lbrace \mathcal{J}_n, \mathcal{J}_{\mathrm{HR}} \rbrace$, in which each pair consists of a noisy LR image and a clean HR image.
\vspace{-4mm}
\paragraph{Baselines}
For performing the noisy image SR using the generated paired dataset, we trained an EDSR-baseline~\cite{edsr} model which contains $16$ residual blocks and $1.5$M parameters. To stabilize the training procedure, we followed the training strategy   in~\cite{edsr}. We pre-trained an EDSR-baseline (EDSR-b) model with clean LR and HR pairs $\lbrace \mathcal{J}_c, \mathcal{J}_{\mathrm{HR}} \rbrace$. Next, we fine-tuned the pre-trained model with the training data synthesized by SCGAN. The fine-tuned model is our final model. 

We regard the performance of the model (EDSR-b) previously trained on clean LR and HR pairs as the { lower bound}. Besides, we also trained an EDSR-baseline model (EDSR*) on images with real Gaussian noise and regarded its performance as the { upper bound}. In addition to noise modeling, the noisy image SR task also can be solved using an alternative two-step method, \ie, image denoising followed by SR. As such, we first use the DnCNN-B model to remove noise from test images, and subsequently, we use the EDSR-b to super-resolve the denoised images. The results of this two-step method (Dn+SR) are shown in Table~\ref{tb:sr-results}. For a fair comparison, the EDSR models in each method are trained with the same hyper-parameters. 

\begin{table}[h]
	\caption{\small{The average PSNR (in dB) results of noisy image SR on the DIV2K and B100 datasets, $\sigma=10$.}}
	\centering
	\scalebox{0.67}{
		\begin{tabular}{c|c|cccccc}
			\toprule
			Dataset	&	Ratio & EDSR-b & Han {\em et al.}~\cite{han2017dictionary} & Dn+SR  & \textbf{SCGAN} & EDSR* (Oracle) \\ \midrule 
			&	$\times$2    & 23.529 &      -                  & 29.555 & \textbf{29.974} & 30.795 \\ 
			DIV2K				&	$\times$ 3    & 22.691 &      -                  & 27.499 & \textbf{27.931} & 28.298  \\ 
			&	$\times$ 4    & 21.889 &      -                  & 26.212 & \textbf{26.583} & 26.883    \\ \midrule
			&	$\times$ 2   & 24.241 &      27.29              & 28.469 & \textbf{28.501} & 28.772   \\ 
			B100				&	$\times$ 3    & 23.317 &      25.64              & 26.616 & \textbf{26.846} & 27.010   \\ 
			&	$\times$ 4    & 22.592 &       24.74             & 25.544 & \textbf{25.870} & 26.029   \\ \bottomrule
		\end{tabular}
	}
	\label{tb:sr-results}
\end{table}

\vspace{-4mm}
\paragraph{Results}
From the results in Table~\ref{tb:sr-results}, we observe that the performances of SCGAN are better than the state-of-the-art method~\cite{han2017dictionary} and close to the upper bound which is derived from a model trained in a fully supervised way. Besides,  compared with the Dn+SR two-step method, the SCGAN not only has better quantitative PSNR result, but also provides a result with better visual quality. That is because, as shown in the Fig.~\ref{fig:sr}, any remaining noise or distortion caused in the denoising step will be amplified during SR.

\def\1subfig {.09} 
\def \2Subfig {.28}
\begin{figure*}[t]
	\centering
	\begin{subfigure}{\1subfig \linewidth}
		\centering
		\includegraphics[width=0.9\linewidth]{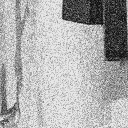}
		\caption*{A}
	\end{subfigure}
	\begin{subfigure}{\2Subfig \linewidth}
		\centering
		\fbox{\begin{minipage}{1 \linewidth}
				\begin{subfigure}{0.32 \linewidth}
					\centering
					\includegraphics[width=0.9\linewidth]{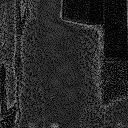}
				\end{subfigure}
				\begin{subfigure}{0.32\linewidth}
					\centering
					\includegraphics[width=0.9\linewidth]{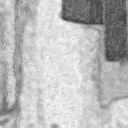}
				\end{subfigure}
				\begin{subfigure}{0.32 \linewidth}
					\centering
					\includegraphics[width=0.9\linewidth]{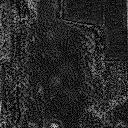}
				\end{subfigure}
		\end{minipage}}
		\caption{Net-1}
		\label{fig:ab-a-net1}
	\end{subfigure}
	\hspace{0.2cm}
	\begin{subfigure}{\2Subfig \linewidth}
		\centering
		\fbox{\begin{minipage}{1 \linewidth}
				\begin{subfigure}{0.32 \linewidth}
					\centering
					\includegraphics[width=0.9\linewidth]{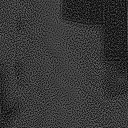}
				\end{subfigure}
				\begin{subfigure}{0.32\linewidth}
					\centering
					\includegraphics[width=0.9\linewidth]{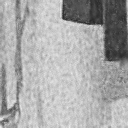}
				\end{subfigure}
				\begin{subfigure}{0.32 \linewidth}
					\centering
					\includegraphics[width=0.9\linewidth]{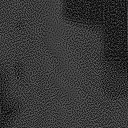}
				\end{subfigure}
		\end{minipage}}
		\caption{Net-2}
		\label{fig:ab-a-net2}
	\end{subfigure}
	\hspace{0.2cm}
	\begin{subfigure}{\2Subfig \linewidth}
		\centering
		\fbox{\begin{minipage}{1 \linewidth}
				\begin{subfigure}{0.32 \linewidth}
					\centering
					\includegraphics[width=0.9\linewidth]{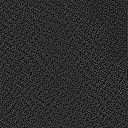}
				\end{subfigure}
				\begin{subfigure}{0.32\linewidth}
					\centering
					\includegraphics[width=0.9\linewidth]{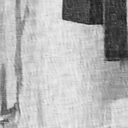}
				\end{subfigure}
				\begin{subfigure}{0.32 \linewidth}
					\centering
					\includegraphics[width=0.9\linewidth]{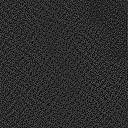}
				\end{subfigure}
		\end{minipage}}
		\caption{Net-3}
		\label{fig:ab-a-net3}
	\end{subfigure}
	
	\begin{subfigure}{\1subfig \linewidth}
		\centering
		\includegraphics[width=0.9\linewidth]{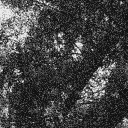}
		\caption*{B}
	\end{subfigure}
	\begin{subfigure}{\2Subfig \linewidth}
		\centering
		\fbox{\begin{minipage}{1 \linewidth}
				\begin{subfigure}{0.32 \linewidth}
					\centering
					\includegraphics[width=0.9\linewidth]{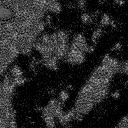}
				\end{subfigure}
				\begin{subfigure}{0.32\linewidth}
					\centering
					\includegraphics[width=0.9\linewidth]{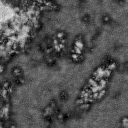}
				\end{subfigure}
				\begin{subfigure}{0.32 \linewidth}
					\centering
					\includegraphics[width=0.9\linewidth]{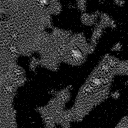}
				\end{subfigure}
		\end{minipage}}
		\caption{Net-1}
		\label{fig:ab-b-net1}
	\end{subfigure}
	\hspace{0.2cm}
	\begin{subfigure}{\2Subfig \linewidth}
		\centering
		\fbox{\begin{minipage}{1 \linewidth}
				\begin{subfigure}{0.32 \linewidth}
					\centering
					\includegraphics[width=0.9\linewidth]{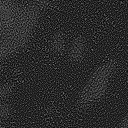}
				\end{subfigure}
				\begin{subfigure}{0.32\linewidth}
					\centering
					\includegraphics[width=0.9\linewidth]{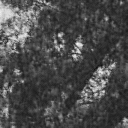}
				\end{subfigure}
				\begin{subfigure}{0.32 \linewidth}
					\centering
					\includegraphics[width=0.9\linewidth]{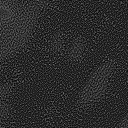}
				\end{subfigure}
		\end{minipage}}
		\caption{Net-2}
		\label{fig:ab-b-net2}
	\end{subfigure}
	\hspace{0.2cm}
	\begin{subfigure}{\2Subfig \linewidth}
		\centering
		\fbox{\begin{minipage}{1 \linewidth}
				\begin{subfigure}{0.32 \linewidth}
					\centering
					\includegraphics[width=0.9\linewidth]{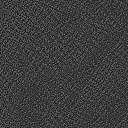}
				\end{subfigure}
				\begin{subfigure}{0.32\linewidth}
					\centering
					\includegraphics[width=0.9\linewidth]{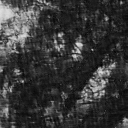}
				\end{subfigure}
				\begin{subfigure}{0.32 \linewidth}
					\centering
					\includegraphics[width=0.9\linewidth]{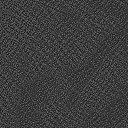}
				\end{subfigure}
		\end{minipage}}
		\caption{Net-3}
		\label{fig:ab-b-net3}
	\end{subfigure}
	
	\caption{Given two noisy patches $A$ and $B$, the outputs of Net-1, Net-2 and Net-3 are shown above. In each bounding box, if the noisy input patch is  $A$, from left to right, there are the extracted noise map ${G(A)}$ of noisy patch $A$, the clean estimation patch ${A-G(A)}$ and $ {G(G(A))}$ which is the output of $G$ taking the extracted noise map as input again.}
	\label{fig:ablation}
\end{figure*}

\subsection{Ablation Study} \label{sec:analysis}

In this subsection, we conduct image blind denoising experiments to investigate effects of the three proposed self-consistent losses, namely the clean consistency $\mathcal{L}_{\mathrm{clean}}$, the pure noise consistency $\mathcal{L}_{\mathrm{pn}}$ and the reconstruction consistency $\mathcal{L}_{\mathrm{rec}}$ in  Eqns.~\eqref{eq:clean}--\eqref{eq:rec}.  We compare the performance of three variants of SCGAN, \ie, Net-1, Net-2, and Net-3, that use different self-consistent constraints. The unpaired image set used for this ablation study is the same as the dataset constructed from the DIV2K for the blind denoising experiment(the second setting).

\begin{table}[h]
	\caption{Three noise modeling networks trained with different combinations of the proposed losses.}
	\vspace{-1em}
	\begin{center}
		\scalebox{0.75}{
			\begin{tabular}{l|l|l}
				\toprule
				Model 	&Losses & 	Description \\
				\midrule
				Net-1 &$\mathcal{L}_{GAN}$ & Only adversarial loss\\
				Net-2 &$ \mathcal{L}_{GAN}$, $\mathcal{L}_{clean} $, $\mathcal{L}_{pn}$	& Net-1 + PureNoise loss + Clean loss\\
				Net-3 &$ \mathcal{L}_{GAN}$, $\mathcal{L}_{clean}$, $\mathcal{L}_{pn}$, $\mathcal{L}_{rec}$	& Net-2 + Rec loss, \textbf{SCGAN }model\\
				\bottomrule
			\end{tabular}
		}
	\end{center}
	\label{tb:ablation}
	\vspace{-2em}
\end{table}

\textbf{Net-1} denotes the model trained only with the adversarial  (GAN) loss. Because of the lack of paired data, the training procedure is severely under-constrained. As shown In Figs.~\ref{fig:ab-a-net1} and~\ref{fig:ab-b-net1}, the noise maps extracted from patches A and B are over-contrast. This leads to low-contrast and even distortion in the ``estimated clean images''. The output of Net-1 is highly correlated to the background information of a given noisy patch. To ameliorate this problem, we added $\mathcal{L}_{\mathrm{clean}}$ and $\mathcal{L}_{\mathrm{pn}}$ to the objective function and trained a SCGAN model  which is now denoted as \textbf{Net-2}. The $\mathcal{L}_{\mathrm{clean}}$ term prevents the model from wrongly extracting textures and backgrounds as noise from clean images. The introduction of the $\mathcal{L}_{\mathrm{pn}} $ term to the objective function ensures that the model extracts objects that can be kept consistent even after being processed by $G$. Compared to the noise maps extracted by Net-1 and Net-2, as shown in Figs.~\ref{fig:ab-a-net1} and~\ref{fig:ab-a-net2}, these two losses ensure that the output noise is distributed more uniformly across the whole global patch. However, the extract noises are still affected by the architecture that exists in the original noisy images. \textbf{Net-3} is trained with a combination of all the three terms of losses. The $\mathcal{L}_{\mathrm{rec}}$ loss helps the generator remove the influence that stems from the architectures in the noisy input. Comparing the noise maps in Figs.~\ref{fig:ab-a-net2} and~\ref{fig:ab-a-net3}, it is easily seen that the noises in the extracted maps show no similarities to the architectures in the original noisy inputs.

For our proposed two-step pipeline, the noise modeling network in the first step extracts the noise map from a given noisy image and generates an estimated clean image. Natural questions arise from our method: for the blind denoising or rain streak removal tasks, it appears to be more natural to use the estimated clean image as the final de-noised output. Why did we not do so? Besides, for the noisy image SR task, it seems natural to generate the clean HR images by regarding the estimated clean images as the inputs to a well-trained EDSR.

To address the first question above, for the blind denoising and rain streak removal tasks, noise in some pixels could be missed by the noise modeling network. Hence, the noise map extracted may be imperfect, and the estimated clean image could still contain noise at these pixels. However, by adding extracted noise maps to clean images, we have a paired training set. When training a deep model for denoising or rain streak removal, in each iteration, we randomly crop paired patches from the generated training set, and the cropped patches contain noise with different intensities. This ensures that the obtained model is able to handle noisy images with various levels of noise, and yields a better de-noised image. To address the second question above, for the noisy image SR  task, our experiments show that the Dn+SR method may result in noise remaining or the loss of details in the denoising step. Both these deficiencies worsen the performance of noisy SR. However, in our method, we train a model for noisy SR in an end-to-end manner. The model directly maps a noisy LR input to its clean HR version, and this ameliorates the aforementioned deficiencies.

\section{Conclusion}
In this paper, we proposed a new unsupervised noise modeling model, \ie, SCGAN,  to extract noise maps from images with unknown noise statistics. To facilitate the model training and improve its  performance, we introduced three self-consistent losses based on intrinsic properties of a noise-modeling  model and noise maps.  We also provided an effective training strategy. Through extensive experiments, we demonstrate the proposed SCGAN   effectively extracted noise maps from noisy images containing  various noise types.  We applied the proposed SCGAN to perform the blind denoising, the rain streak removal, and the noisy image SR tasks, showing its broad applicability. For all the tasks, the SCGAN achieves excellent performances that are close to the performances of models trained in a fully supervised way.

{\small
\bibliographystyle{ieee}
\bibliography{egbib}
}

\section{Appendix}

In the realm of medical image processing, the visual quality of medical images affects its reading and even doctors' diagnosis. Medical images that are corrupted by noise complicate the judgement of patients' condition. However, the acquisition and transmission of medical images inevitably induce some degree of noise. Reducing noise during image acquisition is usually at the cost of longer scanning time or higher radiation dose, which is not a desirable option. Denoising is thus a significant preprocssing step to mitigates the effect of noise.

Existing deep learning methods for medical image denoising generally require a considerable amount of clean and noisy image pairs for model training. Obtaining the aforementioned dataset can be a huge challenge. This emphasizes the necessity of the adoption of unsupervised learning in this domain. 

To this end, we demonstrate the effectiveness of our proposed unsupervised SCGAN in denoising of cephalometric X-ray images. This additional experiment is also an example of generalizing the SCGAN to other domain, not limited to natural images.

\subsection{Experiments}

\noindent\textbf{Dataset} We evaluate the proposed SCGAN model on A dental radiography dataset~\citelatex{wang2016benchmark}, which consists of 400 cephalometric X-ray images with $1935\times 2400$ resolution. Noise corruption is simulated with injection of Gaussian noise to clean 8-bit gray-scale images. We set Gaussian noise to zero mean and standard deviation of 25. We randomly held 100 images out from the dataset as the test set. The remaining 300 images are further split into clean image set $\mathcal{J}_c$ and noise image set $\mathcal{I}_n$. Each set has 150 images.\\

\noindent\textbf{Baselines} Popular medical image denosing methods, i.e., wavelet~\citelatex{chang2000adaptive},  total variance minimization (TV)~\citelatex{sidky2008image} and BM3D~\citelatex{dabov2006image}, as well as start-of-the-art unsupervised denoising method Noise2self~\citelatex{batson2019noise2self} are evaluated on the same set of data for comparison.

\noindent\textbf{Results} Quantitative comparison are shown in Table~\ref{tab:appendix}. The average PSNR results reveals the fact that all methods are able to suppress the corruption by Gaussian noise to cephalometric X-ray image. Deep-learning-based methods (Noise2self and SCGAN)  outperforms traditional methods (wavelet and TV) by a large margin. Even though BM3D archives 31.90dB in average PNSR, surpassing that of our SCAN, it worths to note that BM3D is a non-blind method which relies on paired images. For fair comparison, we trained SCGAN with paired images and mark this version as SCGAN*. SCGAN* obtained higher PSNR compared to BM3D. Among all blind methods, Noise2self achieves highest PSNR and SCGAN follows. 

\begin{figure*}[h]
	\centering
	\begin{subfigure}{\1subfig \linewidth}
		\centering
		\includegraphics[width=\2subfig 
		\linewidth]{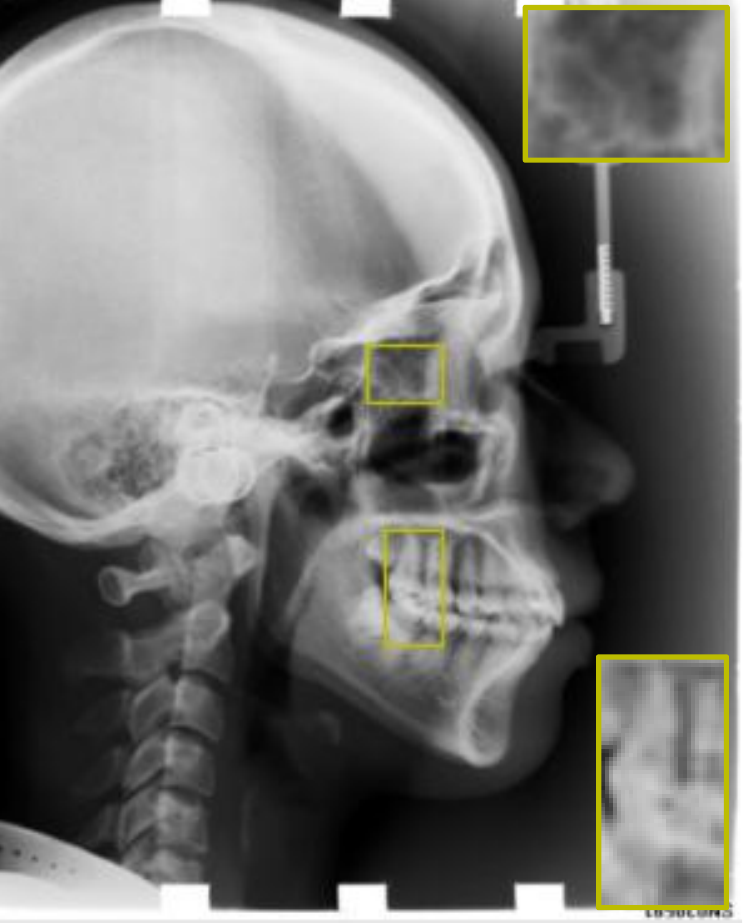}
    \end{subfigure}
	\begin{subfigure}{\1subfig \linewidth}
		\centering
		\includegraphics[width=\2subfig \linewidth]{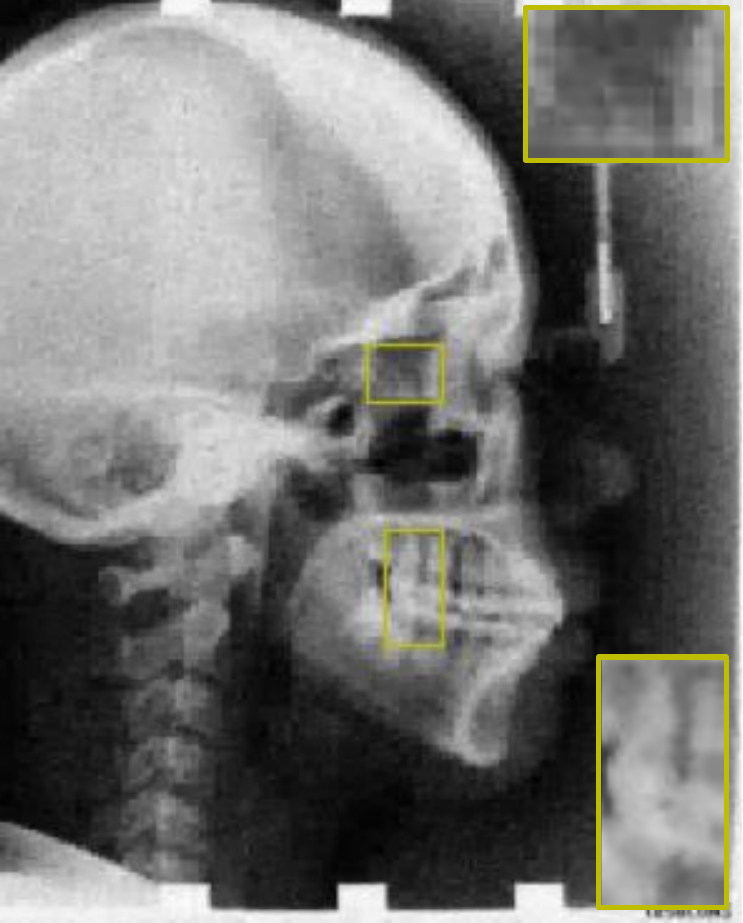}
    \end{subfigure}
    \begin{subfigure}{\1subfig \linewidth}
		\centering
		\includegraphics[width=\2subfig \linewidth]{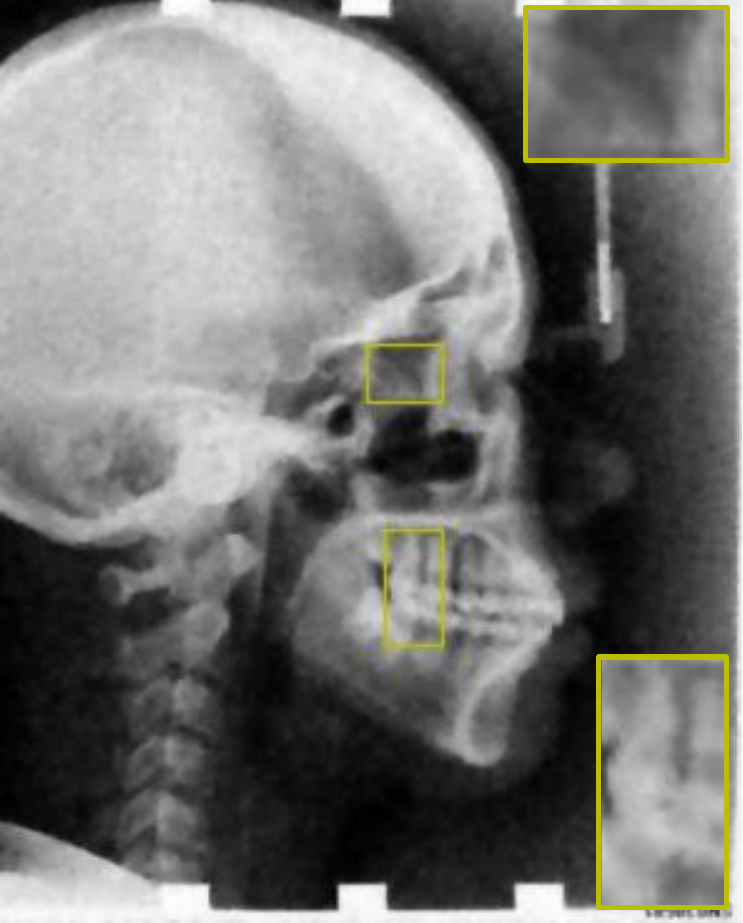}
    \end{subfigure}
    \begin{subfigure}{\1subfig \linewidth}
		\centering
		\includegraphics[width=\2subfig \linewidth]{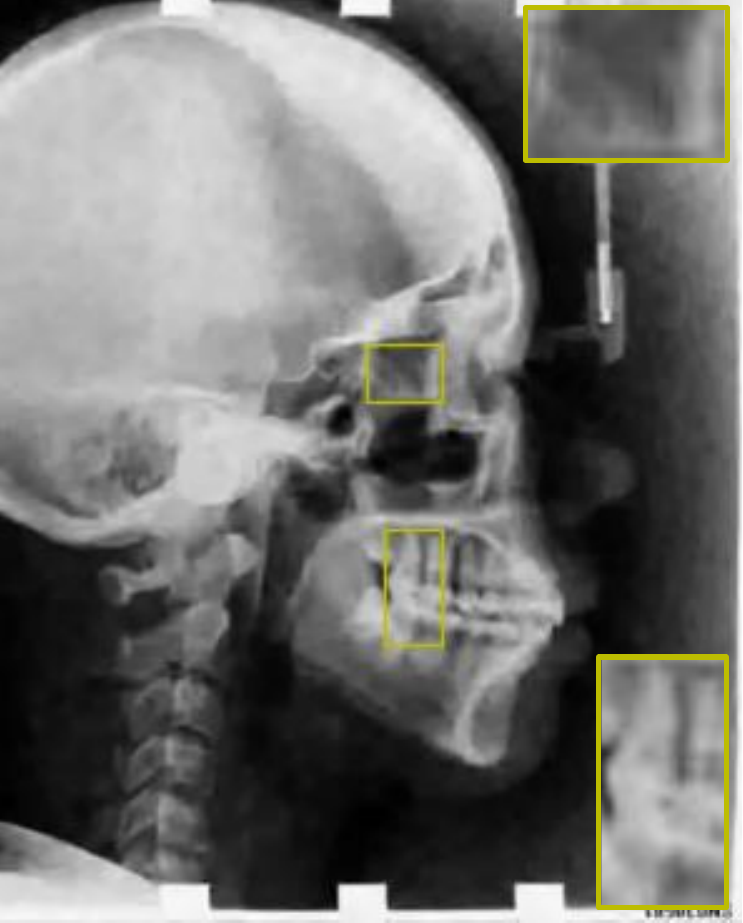}
    \end{subfigure}
    \begin{subfigure}{\1subfig \linewidth}
		\centering
		\includegraphics[width=\2subfig \linewidth]{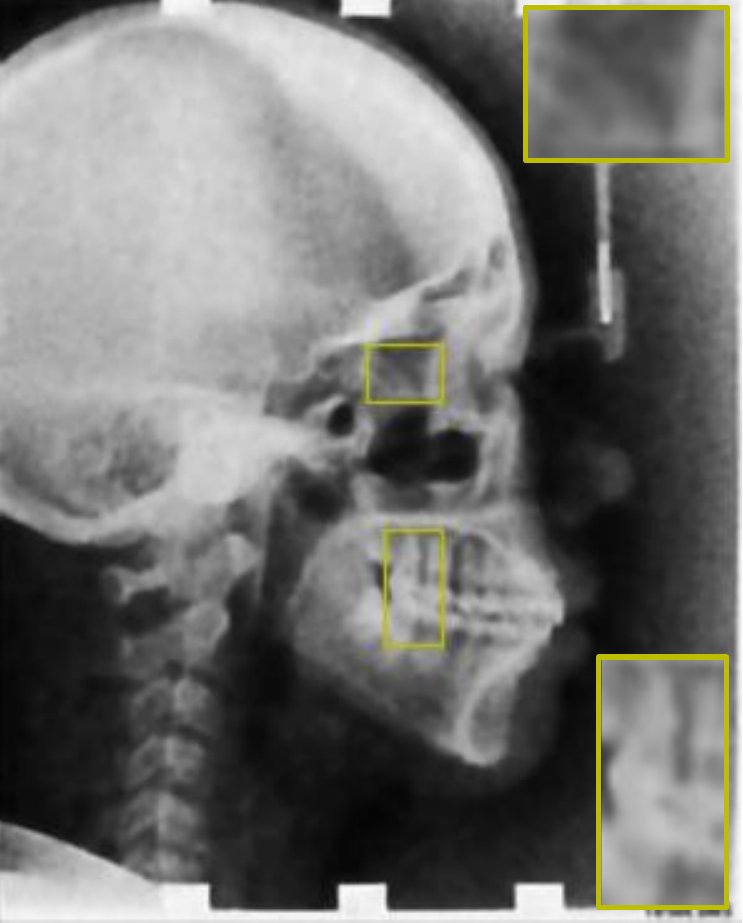}
    \end{subfigure}
    \begin{subfigure}{\1subfig \linewidth}
		\centering
		\includegraphics[width=\2subfig \linewidth]{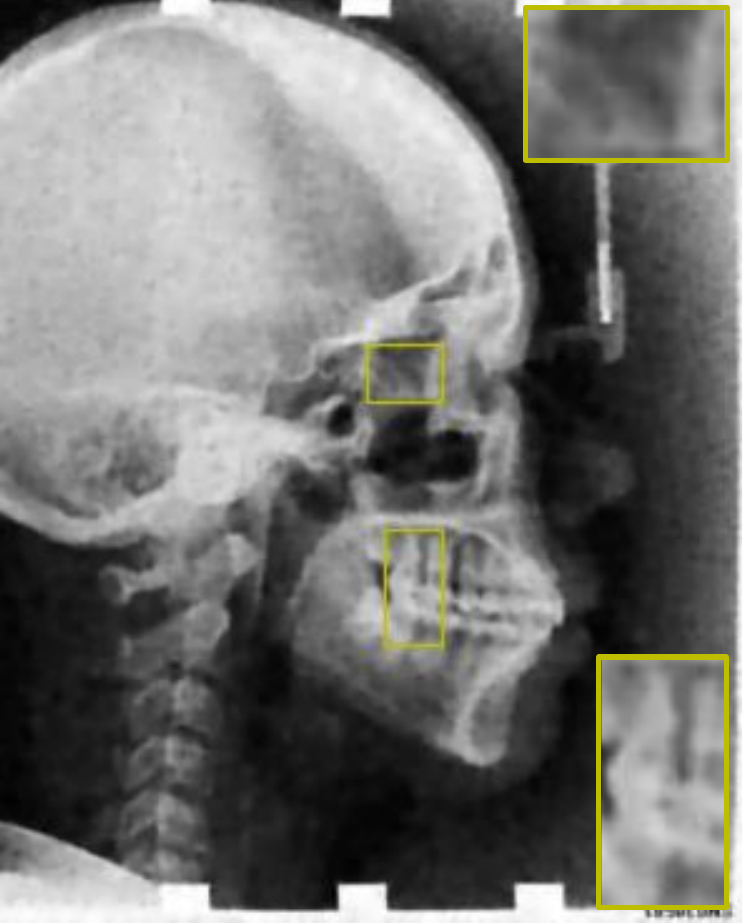}
    \end{subfigure}
    \begin{subfigure}{\1subfig \linewidth}
		\centering
		\includegraphics[width=\2subfig \linewidth]{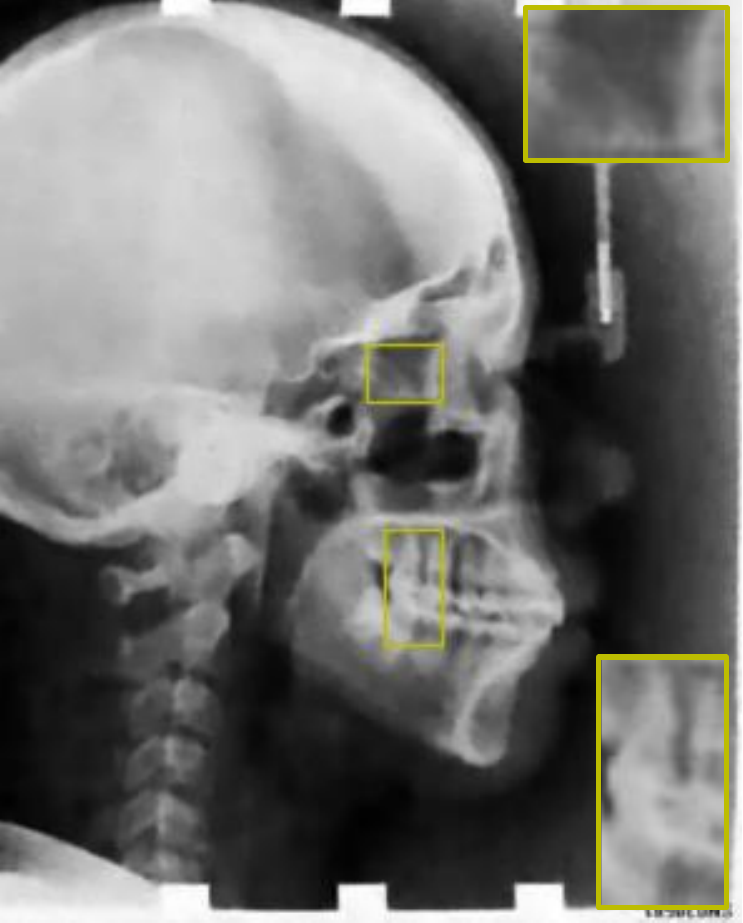}
    \end{subfigure}
    \vspace{2mm}
    \begin{subfigure}{\1subfig \linewidth}
		\centering
		\includegraphics[width=\2subfig \linewidth]{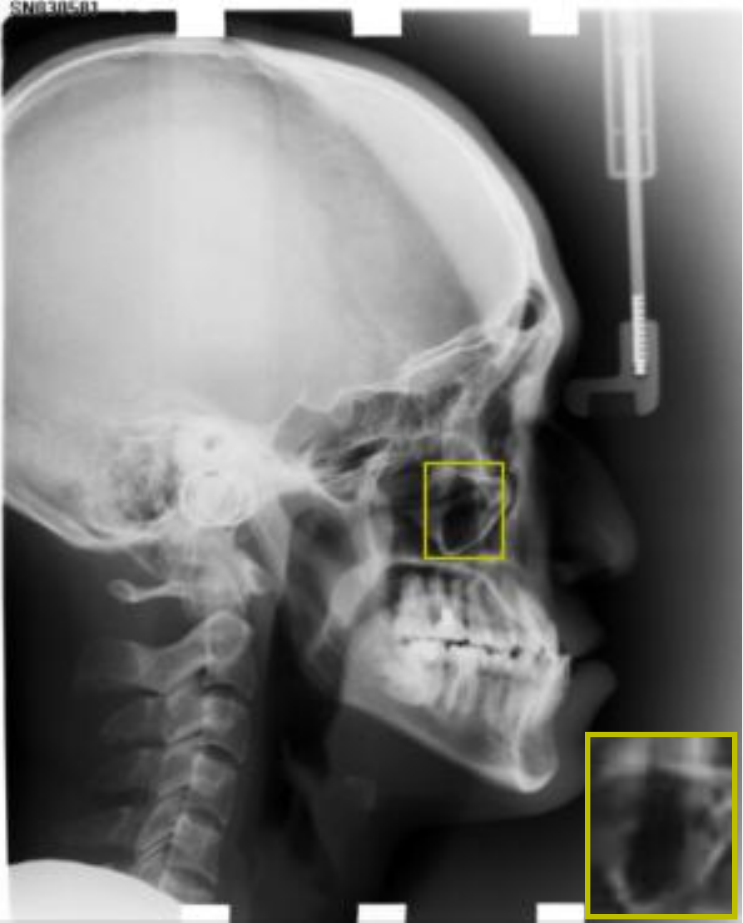}
		\caption{Ground-Truth}
    \end{subfigure}
    \begin{subfigure}{\1subfig \linewidth}
		\centering
		\includegraphics[width=\2subfig \linewidth]{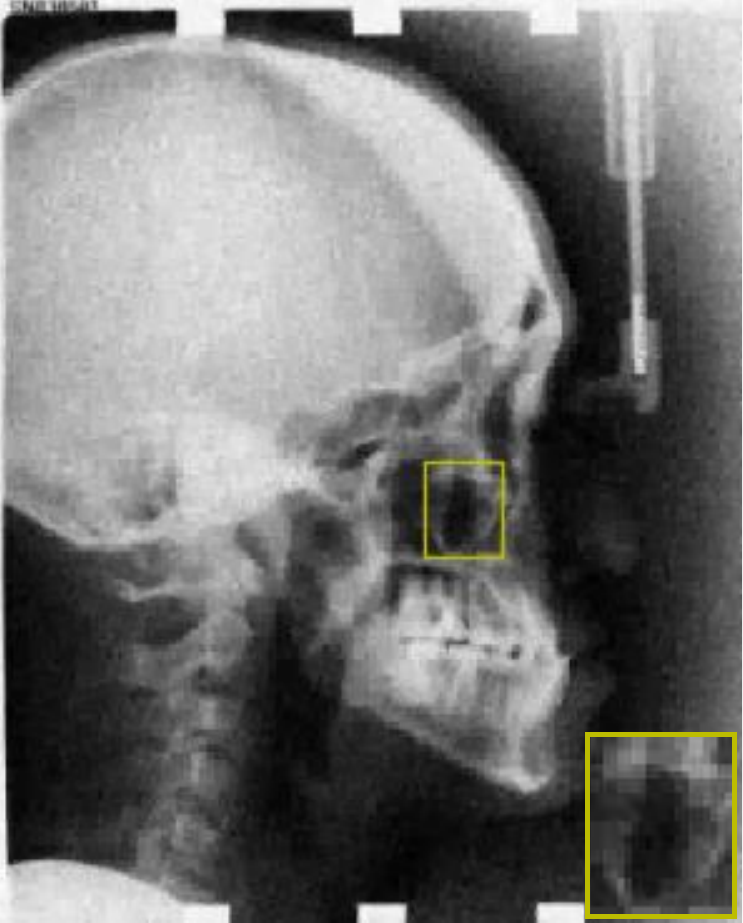}
		\caption{Wavelet}
    \end{subfigure}
    \begin{subfigure}{\1subfig \linewidth}
		\centering
		\includegraphics[width=\2subfig \linewidth]{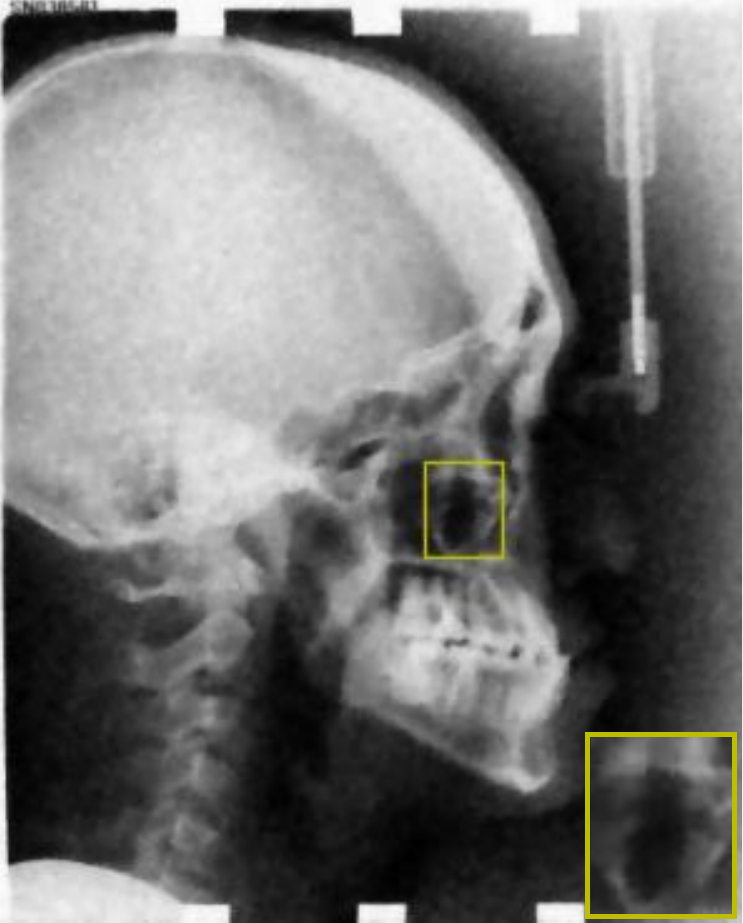}
		\caption{TV}
    \end{subfigure}
    \begin{subfigure}{\1subfig \linewidth}
		\centering
		\includegraphics[width=\2subfig \linewidth]{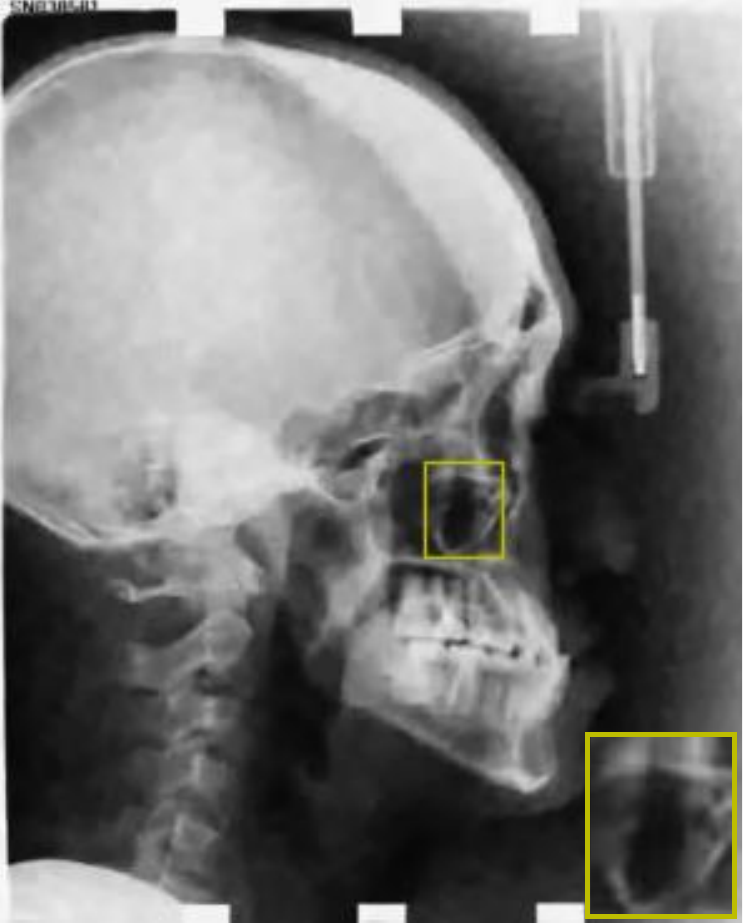}
		\caption{BM3D}
    \end{subfigure}
    \begin{subfigure}{\1subfig \linewidth}
		\centering
		\includegraphics[width=\2subfig \linewidth]{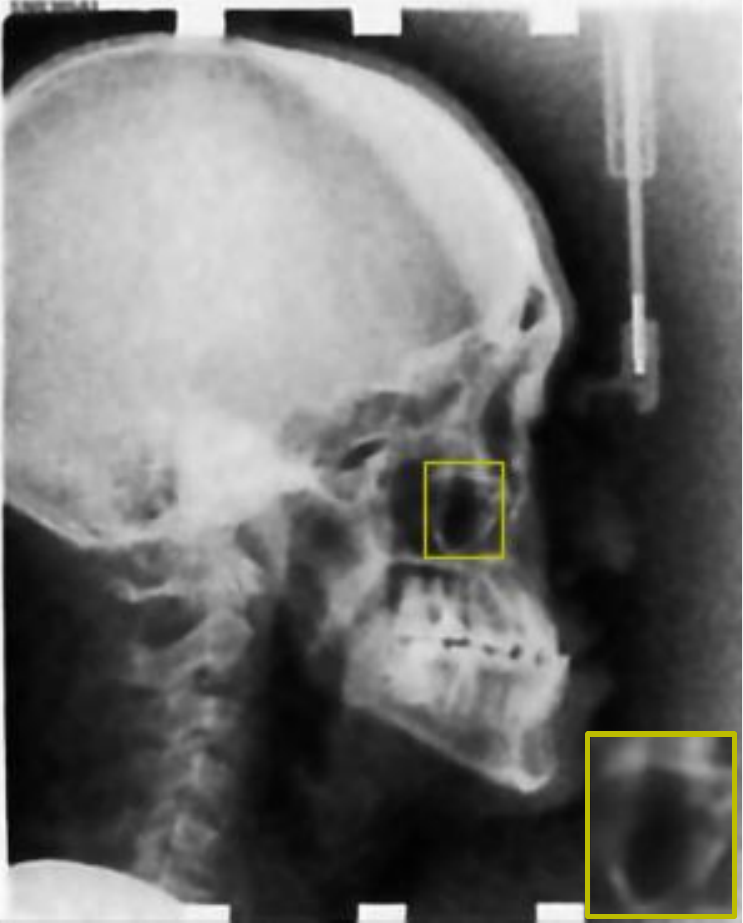}
		\caption{Noise2self}
    \end{subfigure}
    \begin{subfigure}{\1subfig \linewidth}
		\centering
		\includegraphics[width=\2subfig \linewidth]{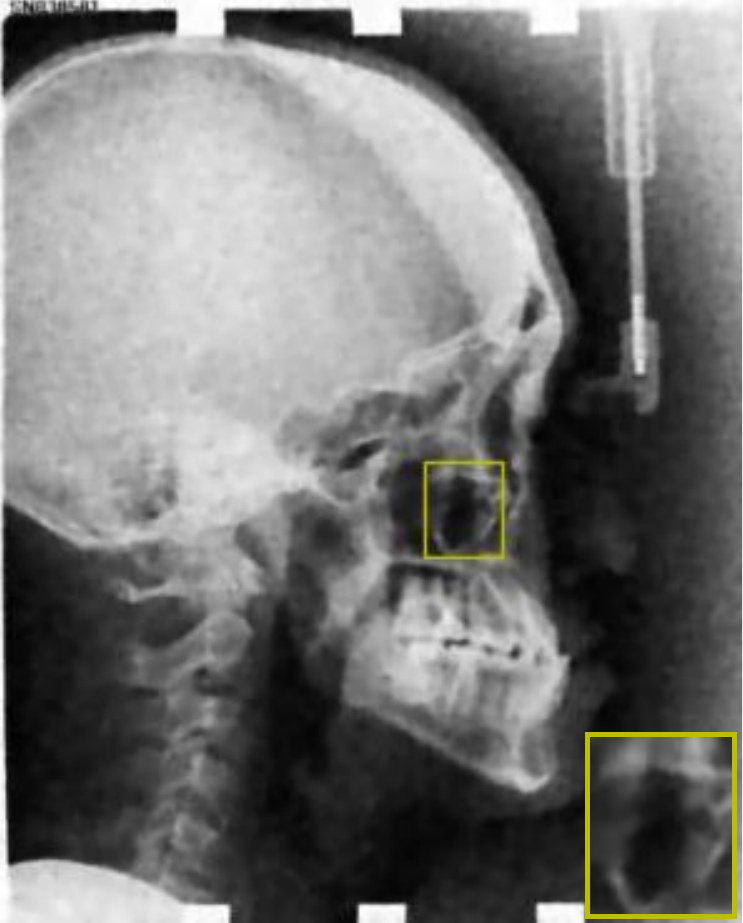}
		\caption{SCGAN}
    \end{subfigure}
    \begin{subfigure}{\1subfig \linewidth}
		\centering
		\includegraphics[width=\2subfig \linewidth]{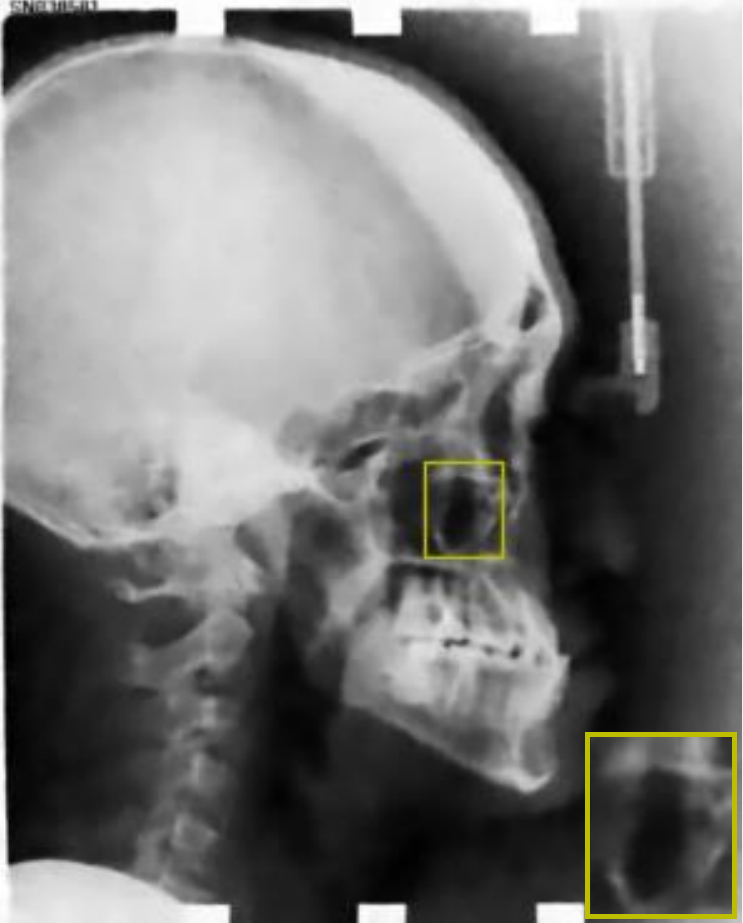}
		\caption{SCGAN*}
    \end{subfigure}
    
    \caption{Comparison of denosing performance of different methods for the image of `325' and the image of `326' from the dental radiography dataset. SCGAN* stands for SCGAN trained in paired manner.}\label{fig:appendix}
\end{figure*}

Figure~\ref{fig:appendix} shows qualitative comparison among all methods. Obviously, fine details are difficult to fully recover from noise corruption. Wavelet denoising leads to blurriest and the most pixelated images. The rest methods produce images with similar visual quality. However, when we take a close look at the regions of interest as indicated in Figure~\ref{fig:appendix}, SCGAN is the best at preserving edges and fine details, even when compared to Noise2self. 

We hence conclude that SCGAN is effective in denoising of medical images. Besides, the proposed SCGAN shows its superiority in preserving fine details, which can help avoid the loss of information in the pre-processing steps of medical images.

The demo codes for this task are released\footnote{\url{https://www.dropbox.com/s/3l59pmn55qq71go/SCGAN_task_cephalometric.zip?dl=0}}.

\begin{table}[h]
\centering
\caption{\small The average PSNR (in dB) results of  cephalometric X-ray images denosing. BM3D is a non-blind method. SCGAN* is the resulting model by training SCGAN in paired manner.}
\scalebox{0.5}{
\begin{tabular}{ l|l l l l l l l }
    \toprule
    Method & Noisy images & Wavelett~\citelatex{chang2000adaptive} & TV~\citelatex{sidky2008image}& BM3D~\citelatex{dabov2006image}& Noise2self~\citelatex{batson2019noise2self}& SCGAN & SCGAN*\\
    \hline 
    PSNR (dB) & 20.56 & 28.57 & 30.67 & 31.90 & 32.06 & 31.43 & 32.29\\
    \bottomrule
\end{tabular}
}
\label{tab:appendix}
\end{table}

{\small
\bibliographystylelatex{ieee}
\bibliographylatex{egbib}
}
\def \1subfig {.13} 
\def \2subfig {1}
\end{document}